\documentclass[10pt,journal]{IEEEtran}

\usepackage{times}
\usepackage{epsfig}
\usepackage{graphicx}
\usepackage{subfigure}
\usepackage{amsmath}
\usepackage{amssymb}
\usepackage[linesnumbered,ruled,vlined]{algorithm2e}
\usepackage{caption}
\usepackage{multirow}
\usepackage{algpseudocode}
\usepackage[normalem]{ulem}
\usepackage{soul}
\usepackage{color}
\usepackage{hyperref}

\ifCLASSOPTIONcompsoc
  \usepackage[nocompress]{cite}
\else
  \usepackage{cite}
\fi
\ifCLASSINFOpdf
\else
\fi
\begin{document}
%
\title{CNN Fixations: An unraveling approach to visualize the discriminative image regions}
%
%
%
%

\author{Konda Reddy Mopuri*,
        Utsav Garg*,
        R. Venkatesh Babu,~\IEEEmembership{Senior Member,~IEEE}
\IEEEcompsocitemizethanks{
\IEEEcompsocthanksitem * denotes equal contribution.
\IEEEcompsocthanksitem Konda Reddy Mopuri (kondamopuri@iisc.ac.in) and R. Venkatesh Babu (venky@iisc.ac.in) are with Video Analytics Lab, Department of Computational and Data Sciences, Indian Institute of Science, Bengaluru, India, 560012. Utsav Garg (utsav002@e.ntu.edu.sg) was an intern at Video Analytics Lab, CDS, IISc, Bangaluru from NTU, Singapore.
}
}

\IEEEtitleabstractindextext{%
\begin{abstract}
Deep convolutional neural networks (CNN) have revolutionized the computer vision research and have seen unprecedented adoption for multiple tasks such as classification, detection, caption generation, etc. However, they offer little transparency into their inner workings and are often treated as black boxes that deliver excellent performance. In this work, we aim at alleviating this opaqueness of CNNs by providing visual explanations for the network's predictions. Our approach can analyze a variety of CNN based models trained for computer vision applications such as object recognition and caption generation. Unlike existing methods, we achieve this via unraveling the forward pass operation. The proposed method exploits feature dependencies across the layer hierarchy and uncovers the discriminative image locations that guide the network's predictions. We name these locations CNN-Fixations, loosely analogous to human eye fixations. Our approach is a generic method that requires no architectural changes, additional training or gradient computation and computes the important image locations (CNN Fixations). We demonstrate through a variety of applications that our approach is able to localize the discriminative image locations across different network architectures, diverse vision tasks and data modalities.
\end{abstract}

\begin{IEEEkeywords}
Explainable AI, CNN visualization, visual explanations, label localization, weakly supervised localization
\end{IEEEkeywords}}

\maketitle

\IEEEdisplaynontitleabstractindextext

%
\IEEEpeerreviewmaketitle

\ifCLASSOPTIONcompsoc
\IEEEraisesectionheading{\section{Introduction}\label{sec:introduction}}
\else
\section{Introduction}
\label{sec:introduction}
\fi

%
%
%
%
\IEEEPARstart{C}onvolutional Neural Networks (CNN) have demonstrated outstanding performance for a multitude of computer vision tasks ranging from recognition and detection to image captioning. CNNs are complex models to design and train. They are non-linear systems that almost always have numerous local minima and are often sensitive to the training parameter settings and initial state. With time, these networks have evolved to have better architectures along with improved regularizers to train them. For example, in case of recognition, from AlexNet~\cite{deepcnn-nips-2012} in $2012$ with $8$ layers and $60M$ parameters, they advanced to ResNets~\cite{resnet-cvpr-2016} in $2015$ with hundreds of layers and $1.7M$ parameters. Though this has resulted in a monotonic increase in performance on many vision tasks (e.g. recognition on ILSVRC~\cite{imagenet-ijcv-2015}, semantic segmentation on PASCAL~\cite{pascal-voc-2007}), the model complexity has increased as well.

\begin{figure}[ht]
\centering
\noindent\begin{minipage}{.48\textwidth}
  \centering
  \includegraphics[width=.32\textwidth]{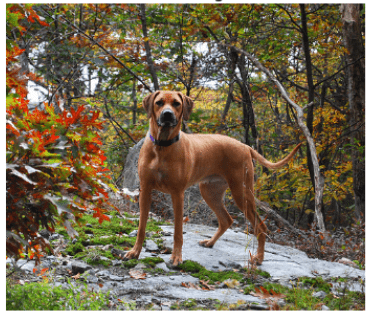}\hfill
  \includegraphics[width=.32\textwidth]{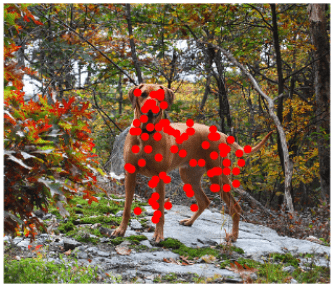}\hfill
  \includegraphics[width=.32\textwidth]{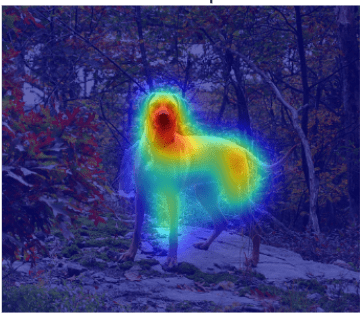}
  \vspace{0.01\textwidth}
\end{minipage}
\noindent\begin{minipage}{.48\textwidth}
  \centering
  \includegraphics[width=.32\textwidth]{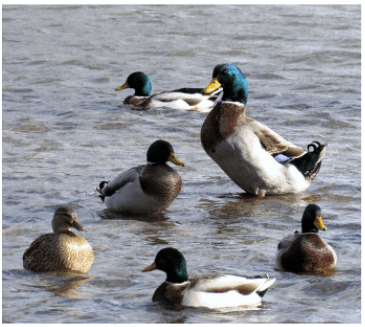}\hfill
  \includegraphics[width=.32\textwidth]{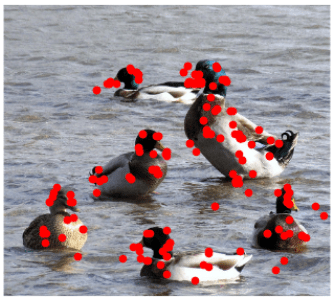}\hfill
  \includegraphics[width=.32\textwidth]{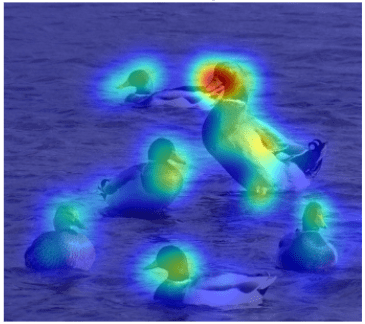}
  \vspace{0.01\textwidth}
\end{minipage}
\caption{CNN fixations computed for a pair of sample images from ILSVRC validation set. Left column: input images. Middle column: corresponding CNN fixations (locations shown in red) overlaid on the image. Right column: The localization map computed form the CNN fixations via Gaussian blurring.
}
\label{fig:sampleMaps}
\end{figure}

In spite of such impressive performance, CNNs continue to be complex machine learning models which offer limited transparency. Current models shed little light on explaining why and how they achieve higher performance and as a result are treated as black boxes. Therefore, it is important to understand what these networks learn in order to gain insights into their representations. One way to understand CNNs is to look at the important image regions that influence their predictions. In cases where the predictions are inaccurate, they should be able to offer visual explanations (as shown in Fig.\ref{fig:wrongPreds}) in terms of the regions responsible for misguiding the CNN. Visualization can play an essential role in understanding CNNs and in devising new design principles (e.g., architecture selection shown in~\cite{deconv-eccv-2014}). With the availability of rich tools for visual exploration of architectures during training and testing, one can reduce the gap between theory and practice by verifying the expected behaviours and exposing the unexpected behaviours that can lead to new insights. 
Towards this, many recent works~(e.g.~\cite{backprop-iclrw-2014,guidedbackprop-iclrw-2015,cam-cvpr-2016,gradcam-iccv-2017,exbp-eccv-2016}) have been proposed to visualize CNNs' predictions. The common goal of these works is to supplement the label predicted by the classifier (CNN) with the discriminative image regions (as shown in Fig.~\ref{fig:compareMaps} and Fig.~\ref{fig:captionMaps}). These maps act as visual explanations for the predicted label and make us understand the class specific patterns learned by the models. Most of these methods utilize the gradient information to visualize the discriminative regions in the input that led to the predicted inference. Some (e.g. \cite{cam-cvpr-2016}) are restricted to work for specific network architectures and output low resolution visualization maps that are interpolated to the original input size.

On the other hand, we propose a visualization approach that exploits the learned feature dependencies between consecutive layers of a CNN using the forward pass operations. That is, in a given layer, for a chosen neuron activation, we can determine the set of positively correlated activations from the previous layer that act as evidence. We perform this process iteratively from the softmax layer till the input layer to determine the discriminative image pixels that support the predicted inference (label). In other words, our approach locates the image regions that were responsible for the CNN's prediction. We name them \emph{CNN fixations}, loosely analogous to the human eye fixations. By giving away these regions, our method makes the CNNs more expressive and transparent by offering the needed visual explanations. We highlight (as shown in Fig.~\ref{fig:sampleMaps}) the discriminative regions by tracing back the corresponding label activation via strong neuron activation paths onto the image plane. Note that we can visualize not only the label activations present in the softmax layer but also any neuron in the model's architecture. Our method offers a high resolution, pixel level localizations. Despite the simplicity of our approach, it could reliably localize objects in case networks trained for recognition task across different input modalities (such as images and sketches) and uncover objects responsible for the predicted caption in case of caption generators (e.g. \cite{SnT-pami-2016}).

The major contributions of this paper can be listed as follows:
	\begin{itemize}
    \item A simple yet powerful method that exploits feature dependencies between a pair of consecutive layers in a CNN to obtain discriminative pixel locations that guide its prediction.
    \item We demonstrate using the proposed approach that CNNs trained for various vision tasks (e.g. recognition, captioning) can reliably localize the objects with little additional computations compared to the gradient based methods.
    \item We show that the approach generalizes across different generations of network architectures and across different data modalities. Furthermore, we demonstrate  the effectiveness of our method through a multitude of applications.
\end{itemize}

Rest of this paper is organized as follows:  section~\ref{sec:relWorks} presents and discusses existing works that are relevant to the proposed method, section \ref{sec:pa} presents the proposed approach in detail, section~\ref{sec:expts} demonstrates the effectiveness of our approach empirically on multiple tasks, modalities and deep architectures, and finally section~\ref{sec:conclu} presents the conclusions.

\section{Related Work}
\label{sec:relWorks}
Our approach draws similarities to recent visualization works. A number of attempts~(e.g.~\cite{relevance-kde-2008,layerwiserelevance-plos-2015, backprop-iclrw-2014,deconv-eccv-2014,guidedbackprop-iclrw-2015,cam-cvpr-2016,gradcam-iccv-2017,exbp-eccv-2016}) have been made in recent time to visualize the classifier decisions and deep learned features. 

Most of these works are gradient based approaches that find out the image regions which can improve the predicted score for a chosen category. Simonyan \textit{et al.}~\cite{backprop-iclrw-2014} measure sensitivity of the classification score for a given class with respect to a small change in pixel values. They compute partial derivative of the score in the pixel space and visualize them as saliency maps. They also show that this is closely related to visualizing using deconvolutions by Zeiler \textit{et al}~\cite{deconv-eccv-2014}. The deconvolution~\cite{deconv-eccv-2014} approach visualizes the features (visual concepts) learned by the neurons across different layers. Guided backprop~\cite{guidedbackprop-iclrw-2015} approach modifies the gradients to improve the visualizations qualitatively.

Zhou \textit{et al.}~\cite{cam-cvpr-2016} showed that class specific activation maps can be obtained by combining the feature maps before the GAP (Global Average Pooling) layer according to the weights connecting the GAP layer to the class activation in the classification layer. However, their method is architecture specific, restricted to networks with GAP layer. Selvaraju \textit{et al.}~\cite{gradcam-iccv-2017} address this issue by making it a more generic approach utilizing gradient information. Despite this, \cite{gradcam-iccv-2017} still computes low resolution maps (e.g. $13 \times 13$). Majority of these methods compute partial derivatives of the class scores with respect to image pixels or intermediate feature maps for localizing the image regions.

Another set of works (e.g.~\cite{relevance-kde-2008,layerwiserelevance-plos-2015,preddiff-iclr-2017,wssc-cvpr-2016}) take a different approach and assign a relevance score for each feature with respect to a class. The underlying idea is to estimate how the prediction changes if a feature is absent. Large difference in prediction indicates that the feature is important for prediction and small changes do not affect the decision. In~\cite{wssc-cvpr-2016}, authors find out the probabilistic contribution of each image patch to the confidence of a classifier and then they incorporate the neighborhood information to improve their weakly supervised saliency prediction.
Zhang \textit{et al.}~\cite{exbp-eccv-2016} compute top down attention maps at different layers in the neural networks via a probabilistic winner takes all framework. They compute marginal winning probabilities for neurons at each layer by exploring feature expectancies. At each layer, the attention map is computed as the sum of these probabilities across the feature maps.

Unlike these existing works, the proposed approach finds the responsible pixel locations by simply unraveling the underlying forward pass operations. Starting from the neuron of interest (e.g. the predicted category label), we rely only on the basic convolution operation to figure out the visual evidence offered by the CNNs. Most of the existing works (e.g. \cite{deconv-eccv-2014, guidedbackprop-iclrw-2015}) realize the discriminative regions via reconstructing the chosen activation. Whereas, our method obtains a binary output at every layer via identifying the relevant neurons. At each layer we can obtain a heat map by simple Gaussian blurring of the binary output. Note that the proposed \textit{CNN-fixations} method has no hyper-parameters or heuristics in the entire process of back tracking the evidence from the softmax layer onto the input image. Fundamentally, our approach exploits the excitatory nature of neurons, which is, being positively correlated and to fire for a specific stimulus (input) from the preceding layer. Though the proposed approach is simple and intuitive in nature, it yields accurate and high resolution visualizations. Unlike majority of the existing works, the proposed method does not require to perform gradient computations, prediction differences, winning probabilities for neurons. Also, the proposed approach poses no architectural constraints and just requires a single forward pass and backtracking operations for the selected neurons that act as the evidence.

\begin{figure*}[th]
    \centering
    \begin{minipage}{.48\textwidth}
        \centering
        \includegraphics[width=\textwidth]{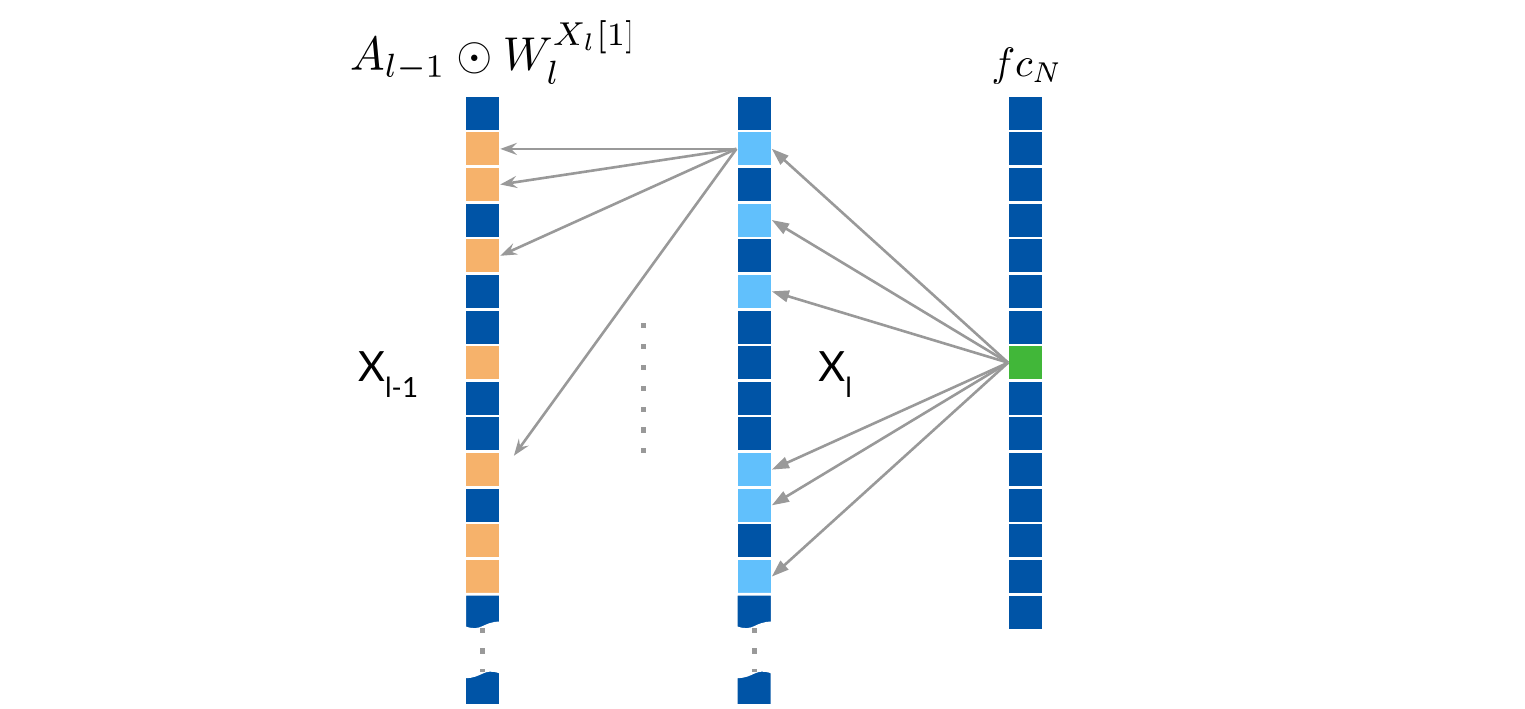}
        \caption{Evidence localization shown between a pair of $fc$ layers. Note that in layer $l-1$, operation is shown for one discriminative location in $X_{l}$. The dark blue color in layers $l-1$ and $l$ indicates locations with $C<0$.}
        \label{fig:fc2fc}        
    \end{minipage}%
    \hfill
    \begin{minipage}{0.48\textwidth}
        \centering
        \includegraphics[width=\textwidth]{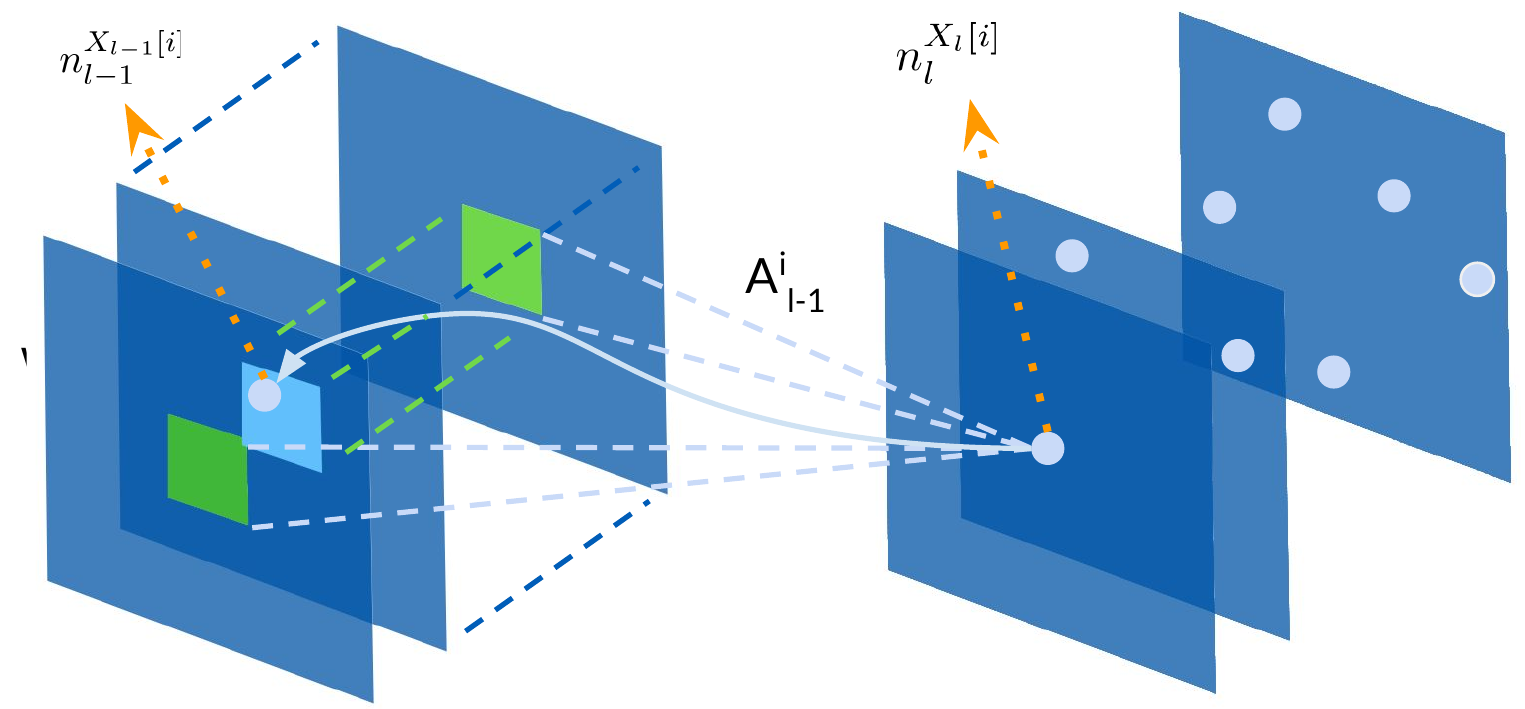}
        \caption{Evidence localization between a pair of convolution layers $l$ and $l-1$. $A_{l-1}^{X_l[i]}$ is the receptive field corresponding to $X_l[i]$. Note that $W_l^{X_l[i]}$ is not shown, however the channel (feature) with maximum contribution (shown in light blue) is determined based on $A_{l-1}^{X_l[i]} \odot W^{X_l[i]}_l $.}
        \label{fig:conv}
    \end{minipage}
\end{figure*}

\section{Proposed Approach}
\label{sec:pa}
In this section, we describe the underlying operations in the proposed approach to determine the discriminative image locations that guide the CNN to its prediction. Note that the objective is to provide visual explanations for the predictions (e.g. labels or captions) in terms of the important pixels in the input image (as shown in Fig.~\ref{fig:sampleMaps} and~\ref{fig:captionMaps}).

Typical deep network architectures have basic building blocks in the form of fully connected, convolution, skip connections and pooling layers or LSTM units in case of captioning networks. In this section we explain our approach for tracing the visual evidence for the prediction across these building blocks onto the image. The following notation is used to explain our approach: we start with a neural network with $N$ layers, thus, the layer indices $l$ range from $1,2, \ldots N$. At layer $l$, we denote the activations as $A_l$ and weights connecting this layer from previous layer as $W_l$. Also, $n_l^k$ represents $k^{th}$ neuron at layer $l$. $X_l$ is the vector of discriminative locations in the feature maps at layer $l$ and $m$ is its cardinality. Note that the proposed approach is typically performed during inference (testing) to provide the visual explanations for the network's prediction.

\subsection{Fully Connected}\label{subsec:fc_text}
A typical CNN for recognition contains a fully connected $(fc)$ layer as the final layer with as many neurons as the number of categories to recognize. During inference, after a forward pass of an image through the CNN, we start with $X_N$ being a vector with one element in the final $fc$ layer, which is the predicted label (shown as green activation in Fig.~\ref{fig:fc2fc}). Note that this can be any chosen activation (neuron) in any layer of the network, our visualization method imposes no restrictions and can localize all the neurons in the architecture.

In case of stacked $fc$ layers, the set $X_{l-1}$ for an $fc$ layer $(l-1)$ will be a vector of indices belonging to important neurons [$n_{l-1}^{X_{l-1}[1]} \ldots n_{l-1}^{X_{l-1}[m]}$] chosen by the succeeding (higher) layer $l$. This set is the list of all neurons in $A_{l-1}$ that contribute to the elements in $X_l$ (in higher layer). That is, for each of the important (discriminative) features determined in the higher layer $(X_l)$, we find the evidence in the current layer $(A_{l-1})$. Thus the proposed approach finds out the evidence by exploiting the feature dependencies between layer $l$ and $l-1$ learned during the training process. We consider all the neurons in $A_{l-1}$ that aid positively (excite) for the task in layer $l$ as its evidence. Algorithm \ref{alg:fc2fc} explains the process of tracing the evidence from a fully connected layer onto the preceding layer.

In case the layer is preceded by a spatial layer (convolution or pooling), we flatten the $3D$ activations $A_{l-1}$ to get a vector for finding the discriminative neurons, finally we convert the indices back to $3D$. Therefore, for spatial layers, $X_{l-1}$ is a list with each entry being three dimensional, namely, \{ \:\:$feature \:(or\:\: channel),\:\: x$, and $y$\}. Figure~\ref{fig:fc2fc} shows how we determine the evidence in the preceding layer for important neurons of an $fc$ layer.

Typically during the evidence tracing, after reaching the first $fc$ layer, a series of conv/pool layers will be encountered. The next subsection describes the process of evidence tracing through a series of convolution layers.

\begin{algorithm}[h]
\SetAlgoLined
\SetKwInOut{Input}{input}  
\Input{$X_{l}$, incoming discriminative locations from higher layer : $\{\:\:X_l[1], \ldots , X_l[m]\:\:\}$  \\
$W_{l}$, weights of higher layer $l$ \\
$A_{l-1}$, activations at current layer $l-1$  
}
\SetKwInOut{Output}{output}  
\Output{$X_{l-1}$, outgoing discriminative locations from the current layer}
$X_{l-1}=\phi$ \\ 
\For{i=1:m }{
    $W_l^{X_l[i]} \gets$ weights of neuron $n_l^{X_l[i]}$\\
    $C \gets$ $A_{l-1} \odot W_l^{X_l[i]}  \:\:\:\:\:// \:\: \text{Hadamard product} $\\ %
    $X_{l-1} \gets$  append ( $X_{l-1}$, $args(C > 0)$ ) 
       } 
 \caption{Discriminative Localization at $fc$ layers.}
 \label{alg:fc2fc}
\end{algorithm}

\subsection{Convolution}\label{subsec:conv_text}
As discussed in the previous subsection, upon reaching a spatial layer, $X_l$ will be a set of $3D$ indices specifying the location of each discriminative neuron. This subsection explains how the proposed approach handles the backtracking between spatial layers. Note that a typical pooling layer will have a $2D$ receptive field and a convolution layer will have a $3D$ receptive field to operate on the previous layer's output. For each important location in $X_l$, we extract the corresponding receptive field activation $A_{l-1}^{X_l[i]}$ in layer $l-1$ (shown as green cuboid in Fig.~\ref{fig:conv}). Hadamard product $(A_{l-1}^{X_l[i]} \odot W^{X_l[i]}_{l} )$ is computed between this receptive activations and the filter weights of  the neuron $n_l^{X_l[i]}$. We then find out the feature (channel) in $A_{l-1}$ that contributes highest (shown in light blue color in Fig.~\ref{fig:conv}) by adding the individual activations from that channel in the hadamard product. That is because, the sum of these terms in the hadamard product gives the contribution of the corresponding feature to excite the discriminative activation in the succeeding layer.

Algorithm \ref{alg:conv} explains this process for convolution layers. In the algorithm, $k_{l-1}$ denotes the kernel size of the convolution filters, ${A_{l-1}}^{X_l[i]}$ are the receptive activations in the previous layer, and hence is a $3D$ spatial blob. Therefore, when the Hadamard product is computed with the weights $(W_l^{X_l[i]})$ of the neuron, the result is also a spatial blob of the same size. We sum the output across $x$ and $y$ directions to locate the most discriminative feature map ``$ch$" (shown in light blue color in Fig.~\ref{fig:conv}). During this transition, spatial location of the activation can also get affected. That means, $(x,y)$ location in the succeeding layer is traced onto the strongest contributing activation of channel ``$ch$" in the current layer. Instead, we can also trace back to the same location within the most contributing channel ``$ch$". However, we empirically found that this is not significantly different from the former alternative. Therefore, in all our experiments, for computational efficiency, we follow the latter alternative of tracking onto the same location as in the succeeding layer. Note that the procedures we follow for evidence tracking across $fc$ (Algo.~\ref{alg:fc2fc}) and $conv$ (Algo.~\ref{alg:conv}) layers are fundamentally similar, except that $conv$ layers operate over $3D$ input blobs, whereas $fc$ layers have a $1D$ input blob (after vectorizing). Algorithm~\ref{alg:conv} explains the process considering the localized input blobs (receptive activations) and the convolution kernels to the exact implementation details.

In case of pooling layers, we extract the $2D$ receptive neurons in the previous layer and find the location with the highest activation. This is because most of the architectures typically use max-pooling to sub-sample the feature maps. The activation in the succeeding layer is the maximum activation present in the corresponding receptive field in the current layer. Thus, when we backtrack an activation across a sub-sampling layer, we locate the maximum activation in its receptive field.

\begin{algorithm}[t]
\SetAlgoLined
\SetKwInOut{Input}{input}  
\Input{$X_{l}$, incoming discriminative locations from higher layer : ${\:\: X_l[1] \ldots X_l[m]\:\:}$  \\
$W_{l}$, weights at layer $l$ \\
$A_{l-1}$, activations at layer $l-1$  
}
\SetKwInOut{Output}{output}  
\Output{$X_{l-1}$, outgoing discriminative locations in the current layer}
$S(.)$: a function that sums a tensor along $xy$ axes \\
$X_{l-1}=\phi$ \\
\For{i=1:m }{
    $W_l^{X_l[i]} \gets \text{weights for neuron }n_l^{X_l[i]}$\\
    $A_{l-1}^{X_l[i]} \gets\text{ receptive activations for neuron }n_l^{X_l[i]}$\\
    $C \gets S(A_{l-1}^{X_l[i]} \odot W_l^{X_l[i]})\://\:\text{Per channel contributions}$\\
    $ch \gets argmax(C)//\:\:\text{Discriminative channel}$\\
    $(P_x,P_y) \gets argmax(A_{l-1}^{X_l[i]} (:,:,ch)) \odot W_l^{X_l[i]} (:,:,ch)) \:\:\: //\:\:\:\text{Discriminative location in channel `ch'}$\\
    $X_{l-1} \gets\text{ append }( X_{l-1} , ch . k_{l-1}^2+P_x . k_{l-1} + P_y )$\\
    }
    $X_{l-1} \gets unique(X_{l-1})$
 \caption{Discriminative Localization at Convolution layers}
 \label{alg:conv}
\end{algorithm}

Thus for a CNN trained for recognition, the proposed approach starts from the predicted label in the final layer and iteratively backtracks through the $fc$ layers and then through the convolution layers onto the image. CNN Fixations (red dots shown in middle column of Fig.~\ref{fig:sampleMaps}) are the final discriminative locations determined on the image. Note that the fixations are $3D$ coordinates since the input image generally contains three channels (R, G and B). However, we consider the union of spatial coordinates ($x$ and $y$) of the fixations neglecting the channel.

\subsection{Advanced architectures: Inception, Skip connections and Densely connected convolutions} 
Inception modules have shown (e.g. Szegedy \textit{et al.}~\cite{googlenet-cvpr-21014}) to learn better representations by extracting multi-level features from the input. They typically comprise of multiple branches which extract features at different scales and concatenate them along the channels at the end. The concatenation of feature maps will have a single spatial resolution but increased depth through multiple scales. Therefore, each channel in `Concat' is contributed by exactly one of these branches. Let us consider an inception layer with activation $A_l$ with $B$ input branches getting concatenated. That means, $A_l$ is concatenation of $B$ outputs obtained via convolving the previous activations $A_{l-1}$ with a set of $B$ different weights $\{W_{lb}\}$ where $b \in \{1,\dots B\}$. For each of the important activations $X_l$ in the inception layer, there is exactly one input branch connecting it to the previous activations $A_{l-1}$. Since we know the number of channels resulted by each of the input branches, we can identify the corresponding input branch for $X_l$ from the channel on which it lies. Once we determine the corresponding input branch, it is equivalent to performing evidence tracing via a $conv$ layer. Hence, we perform the same operations as we perform for a $conv$ layer (discussed in section \ref{subsec:conv_text} and Algorithm~\ref{alg:conv}) after determining which branch caused the given activation.

He \textit{et al.}~\cite{resnet-cvpr-2016} presented a residual learning framework to train very deep neural networks. They introduced the concept of residual blocks (or ResBlocks) to learn residual function with respect to the input. A typical ResBlock contains a skip path and a residual (delta) path. The delta path $(D_{l-1})$ generally consists of multiple convolutional layers and the skip path is an identity connection with no transformation. Ending of the ResBlock performs element wise sum of the incoming skip $(A_{l-1})$ and delta branches $(D_{l-1})$. Note that this is unlike the inception block where each activation is a contribution of a single transformation. Therefore, for each discriminative location in $X_l$, we find the branch (either skip or delta) that has a higher contributing activation and trace the evidence through that route. If for a given location $X_l[i]$, the skip path contributes more to the summation, it is traced directly onto $A_{l-1}$ through the identity transformation. On the other hand, if the delta path contributes more than the skip connection, we trace through the multiple $conv$ layers of the delta path as we explained in section~\ref{subsec:conv_text}. We perform this process iteratively across all the ResBlocks in the architecture to determine the visual explanations. 

Huang~\textit{et al.} introduced Dense Convolutional Networks (DenseNet~\cite{densenet-cvpr-2107}) that connects each layer to every other layer in a feed-forward fashion. For each layer, feature maps of all the earlier layers are used as input and its own feature maps are used as input to later layers. In other words, dense connections can be considered as a combination of skip connections and inception modules. At a given layer $(l)$ in the architecture, a skip path from the previous layer's output $(A_{l-1})$ gets concatenated to the activations (feature maps) computed at this layer $(A_l)$. Note that in case of ResNets, the skip and and delta paths gets added. Therefore, for a given discriminative activation in the current layer, the backtracking has two options: either it belongs to the current feature maps computed at this layer or it is transferred from the previous layer. If it belongs to current set of feature maps, we can backtrack using the conv component (section.~\ref{subsec:conv_text}) of the proposed approach. Else, if it belongs to the feature maps copied from the previous layers, we simply transfer (copy) the discriminative location onto the previous layer, since it is an identity transformation from previous layer to current layer. This process of evidence tracing is performed iteratively till the input layer to obtain the CNN-Fixations. Thus, our method is a generic approach and it can visualize all CNN architectures ranging from the first generation AlexNet~\cite{deepcnn-nips-2012} to the recent DenseNets~\cite{densenet-cvpr-2107}.

\begin{figure*}[th]
\centering

\noindent\begin{minipage}{\textwidth}
  \centering
  \begin{minipage}{.16\textwidth}
  	\centering
    \includegraphics[width=\linewidth]{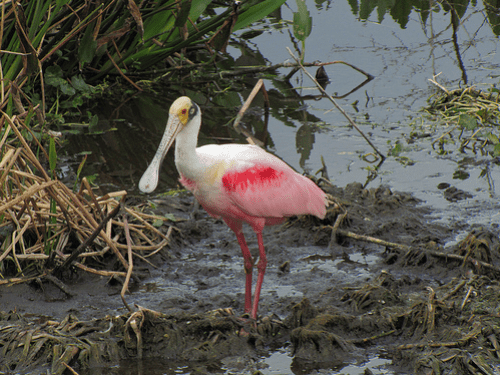}
  \end{minipage}
   \begin{minipage}{.16\textwidth}
   	\centering
    \includegraphics[width=\linewidth]{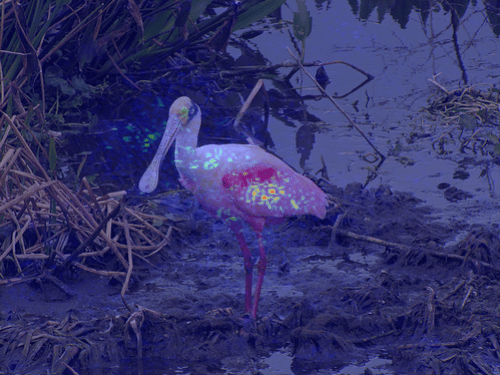}
    
  \end{minipage}
  \begin{minipage}{.16\textwidth}
  	\centering
    \includegraphics[width=\linewidth]{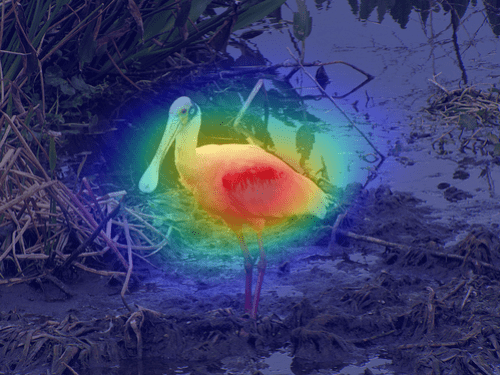}
  \end{minipage}
  \begin{minipage}{.16\textwidth}
  	\centering
    \includegraphics[width=\linewidth]{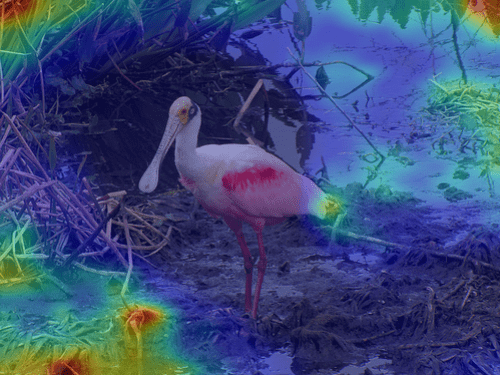}
  \end{minipage}
  \begin{minipage}{.16\textwidth}
  	\centering
    \includegraphics[width=\linewidth]{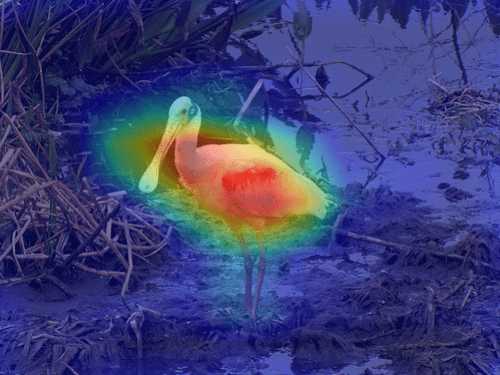}
  \end{minipage}
  \begin{minipage}{.16\textwidth}
  	\centering
    \includegraphics[width=\linewidth]{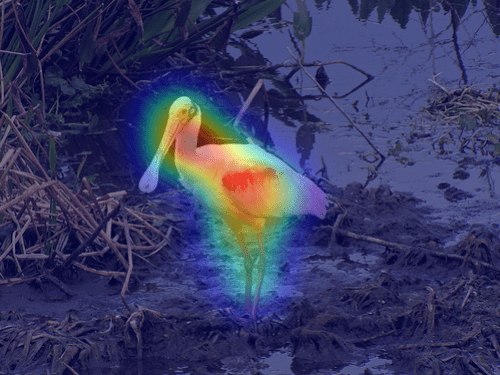}
  \end{minipage}
  \vspace{0.002\textwidth}
\end{minipage}
\noindent\begin{minipage}{\textwidth}
  \centering
  \begin{minipage}{.16\textwidth}
  	\centering
    \includegraphics[width=\linewidth]{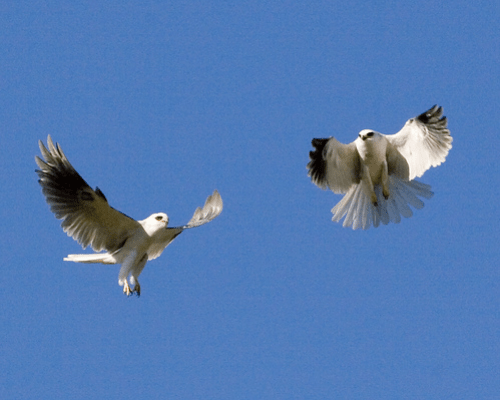}
    \caption*{Input Image}
  \end{minipage}
   \begin{minipage}{.16\textwidth}
   	\centering
    \includegraphics[width=\linewidth]{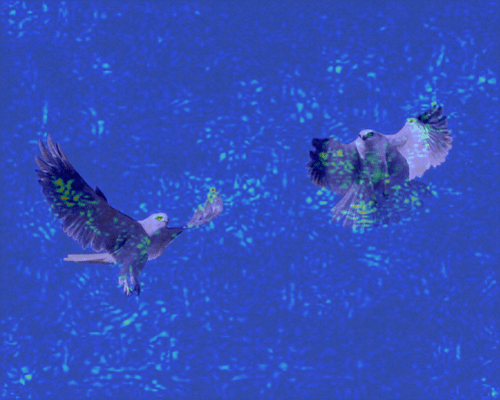}
    \caption*{Backprop~\cite{backprop-iclrw-2014}}
  \end{minipage}
  \begin{minipage}{.16\textwidth}
  	\centering
    \includegraphics[width=\linewidth]{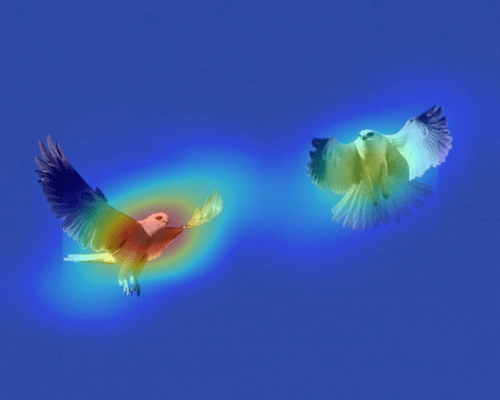}
    \caption*{CAM\cite{cam-cvpr-2016}}
  \end{minipage}
  \begin{minipage}{.16\textwidth}
  	\centering
    \includegraphics[width=\linewidth]{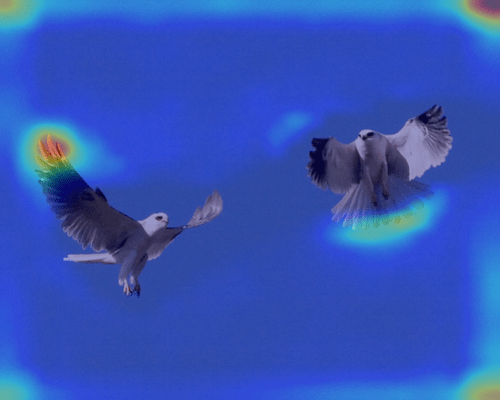}
    \caption*{cMWP~\cite{exbp-eccv-2016}}
  \end{minipage}
  \begin{minipage}{.16\textwidth}
  	\centering
    \includegraphics[width=\linewidth]{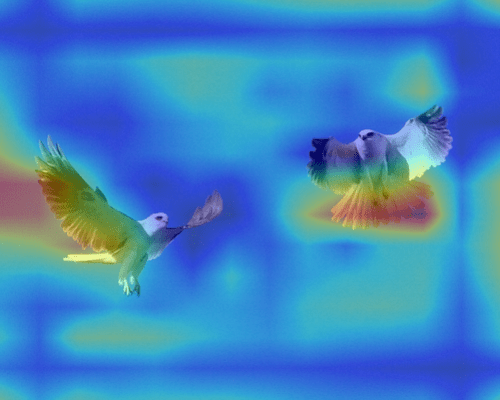}
    \caption*{Grad-CAM~\cite{gradcam-iccv-2017}}
  \end{minipage}
  \begin{minipage}{.16\textwidth}
  	\centering
    \includegraphics[width=\linewidth]{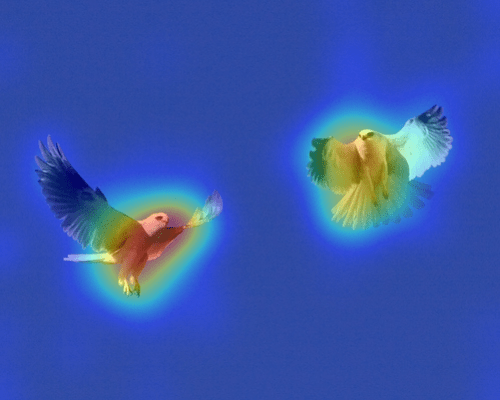}
    \caption*{Proposed}
  \end{minipage}
  \vspace{0.002\textwidth}
\end{minipage}
\caption{Comparison of the localization maps for sample images from ILSVRC validation set across different methods for VGG-16~\cite{vgg-iclr-2015} architecture without any thresholding of maps. For our method, we blur the discriminative image locations using a Gaussian to get the map.}
\label{fig:compareMaps}
\end{figure*}

\begin{figure*}[th]
\centering
\noindent\begin{minipage}{\textwidth}
  \centering
  \begin{minipage}{.16\textwidth}
    \includegraphics[width=\linewidth]{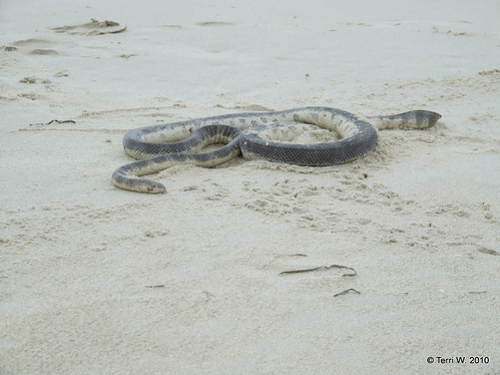}\caption*{Input Image}
  \end{minipage}
   \begin{minipage}{.16\textwidth}
    \includegraphics[width=\linewidth]{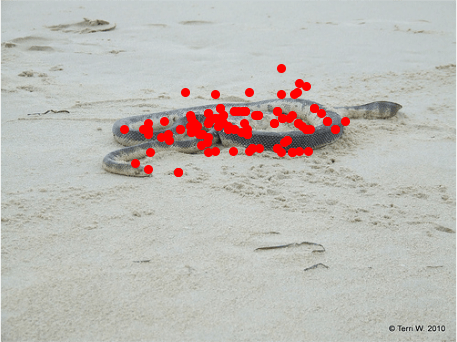}\caption*{Pool1}
  \end{minipage}
  \begin{minipage}{.16\textwidth}
    \includegraphics[width=\linewidth]{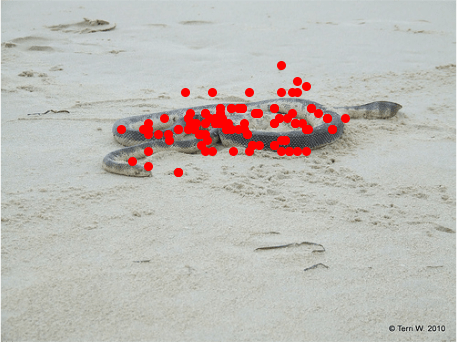}\caption*{Pool2}
  \end{minipage}
  \begin{minipage}{.16\textwidth}
    \includegraphics[width=\linewidth]{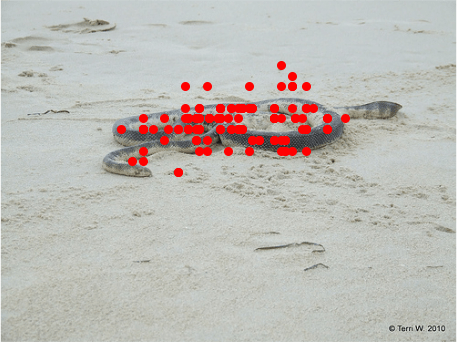}\caption*{Pool3}
  \end{minipage}
  \begin{minipage}{.16\textwidth}
    \includegraphics[width=\linewidth]{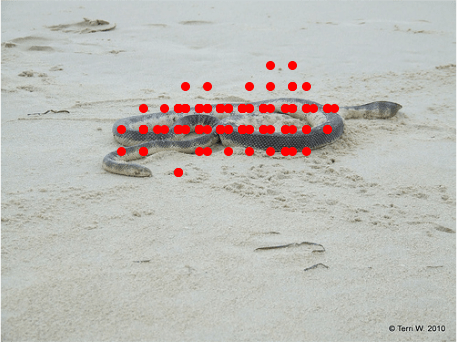}\caption*{Pool4}
  \end{minipage}
  \begin{minipage}{.16\textwidth}
    \includegraphics[width=\linewidth]{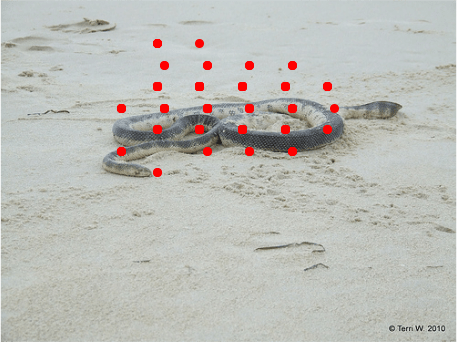}\caption*{Pool5}
  \end{minipage}
  \vspace{0.002\textwidth}
\end{minipage}

\caption{CNN-Fixations at intermediate layers of VGG-16~\cite{vgg-iclr-2015} network. Note that the fixations at deeper layers are also displayed on the original resolution image via interpolating.}
\label{fig:maps-at-intermediate}
\end{figure*}

\subsection{LSTM Units}
\label{sec:lstm}
In this subsection we discuss our approach to backtrack through an LSTM~\cite{lstm-nc-1997} unit used in caption generation networks (e.g. \cite{SnT-pami-2016}). The initial input to the LSTM unit is random state and the image embedding $I$ encoded by a CNN. In the following time steps image embedding is replaced by embedding for the word predicted in the previous time step. An LSTM unit is guided by the following equations~\cite{SnT-pami-2016}:
\begin{equation}\label{eq:l1}
i_t = \sigma(W_{ix}x_t + W_{im}m_{t-1})
\end{equation}
\begin{equation}\label{eq:l2}
f_t = \sigma(W_{fx}x_t + W_{fm}m_{t-1})
\end{equation}
\begin{equation}\label{eq:l3}
o_t = \sigma(W_{ox}x_t + W_{om}m_{t-1})
\end{equation}
\begin{equation}\label{eq:l4}
c_t = f_t \odot c_{t-1} + i_t \odot h(W_{cx}x_t + W_{cm}m_{t-1})
\end{equation}
\begin{equation}\label{eq:l5}
m_t = o_t \odot c_t
\end{equation}
Here, $i$, $f$ and $o$ are the input, forget and output gates respectively of the LSTM and $\sigma$ and $h$ are the sigmoid and hyperbolic-tan non-linearities. $m_t$ is the state of the LSTM which is passed along with the input to the next time step. At each time step, a softmax layer is learned over $m_t$ to output a probability density over a set of dictionary words.

Our approach takes the maximum element in $m$ at the last unrolling and then tracks back the discriminative locations through the four gates individually and then accumulates them as locations on $m_{t-1}$. Tracking back through these gates involves operations similar to the ones discussed in case of fully connected layers \ref{subsec:fc_text}. We iteratively perform backtracking through the time steps till we finally reach the image embedding $I$. Once we reach $I$, we perform the operations discussed in sections \ref{subsec:fc_text} and \ref{subsec:conv_text} to obtain the discriminative locations on the image.
\section{Applications}
\label{sec:expts}
This section demonstrates the effectiveness of the proposed approach across multiple vision tasks and modalities through a variety of applications.

The proposed approach is both network and framework agnostic. It requires no training or modification to the network to get the discriminative locations. The algorithm needs to extract the weights and activations from the network to perform the operations discussed in the sections above. Therefore any network can be visualized with any deep learning framework. For the majority of our experiments we used the Python binding of Caffe~\cite{caffe-acmmm-2014} to access the weights and activations, and we used Tensorflow~\cite{tensorflow2015-whitepaper} in case of captioning networks as the models for Show and Tell~\cite{SnT-pami-2016} are provided in that framework. After finding the important pixels in the image, we perform outlier removal before we compute the heat map. We consider a location to be an outlier, if it is not supported by sufficient neighboring fixations. Specifically, if a fixation has less than certain percentage of total fixations within a given circle around it, we neglect it. Codes for the project are publicly available at \url{https://github.com/val-iisc/cnn-fixations}. Additional qualitative results for some applications are available at \url{http://val.cds.iisc.ac.in/cnn-fixations/}.

\subsection{Weakly Supervised Object Localization}
\label{subsec:objLoc}
We now empirically demonstrate that the proposed CNN fixations approach is capable of efficiently localizing the object recognized by the CNN. Object recognition or classification is the task of predicting an object label for a given image. However, object detection involves not only predicting the object label but also localizing it in the given image with a bounding box. The conventional approach has been to train CNN models separately for the two tasks. Although some works~(e.g.~\cite{fasterrcnn-nips-2015,craft-cvpr-2016}) share features between both tasks, detection models~(e.g.~\cite{fasterrcnn-nips-2015,craft-cvpr-2016,ohem-cvpr-2016}) typically require training data with human annotations of the object bounding
boxes. In this section, we demonstrate that the CNN
models trained to perform object recognition are also capable of localization.

\begin{table}[t]
\centering
\caption{ Error rates for Weakly Supervised Localization of different visualization approaches on ILSVRC validation set. The numbers show error rate, which is $100 - \text{Accuracy of localization}$ (lower the better). (*) denotes modified architecture, bold face is the best performance in the column and underline denotes the second best performance in the column. Note that the numbers are computed for the top-1 recognition prediction.}
\label{tab:objLoc}
\resizebox{0.48\textwidth}{!}{ 
\begin{tabular}{cccccc}
\hline
\textbf{Method} & \textbf{AlexNet} & \textbf{VGG-16} & \textbf{GoogLeNet} & \textbf{ResNet-101} & \textbf{DenseNet-121}\\ \hline
Backprop        & \textbf{65.17}        & 61.12    & 61.31          & 57.97 & 67.49          \\
CAM             & 67.19*           & \uline{57.20*}       & \uline{60.09}              & \textbf{48.34} & \textbf{55.37}          \\
cMWP            & 72.31            & 64.18        & 69.25              & 65.94      &64.97     \\
Grad-CAM        & 71.16            & 56.51        & 74.26              & 64.84 &75.29          \\ \hline
Ours            & \uline{65.70}            & \textbf{55.22}        & \textbf{57.53}              & \uline{54.31}    & \uline{56.72}       \\ \hline
\end{tabular}
}
\end{table}
For our approach, after the forward pass, we backtrack the label on to the image. Unlike other methods our approach finds the important locations (as shown in Figure~\ref{fig:sampleMaps}) instead of a heatmap, therefore we perform outlier removal as follows: we consider a location to be an outlier if the location is not sufficiently supported by neighboring fixation locations. Particularly, if any of the \emph{CNN Fixations} has less than a certain percentage of the fixations present in a given circle around it, we consider it as an outlier and remove it. These two parameters, percentage of points and radius of the circle were found over a held out set, and we found $5\%$ of points and radius equal to $9-11 \%$ of the image diagonal to perform well depending on the architecture. After removing the outliers, we use the best fitting bounding box for the remaining locations as the predicted location for the object. 

We perform localization experiments on the ILSVRC~\cite{imagenet-ijcv-2015} validation set of $50,000$ images with each image having one or multiple objects of a single class. Ground truth consists of object category and the bounding box coordinates for each instance of the objects. Similar to the existing visualization methods (e.g. \cite{cam-cvpr-2016, gradcam-iccv-2017,exbp-eccv-2016}), our evaluation metric is accuracy of localization, which requires to get the prediction correct and obtain a minimum of $0.5$ Intersection over Union (IoU) between the predicted and ground-truth bounding boxes. Table~\ref{tab:objLoc} shows the error rates $(100-\text{Accuracy of localization})$ for corresponding visualization methods across multiple network architectures such as AlexNet~\cite{deepcnn-nips-2012}, VGG~\cite{vgg-iclr-2015}, GoogLeNet~\cite{googlenet-cvpr-21014}, ResNet~\cite{resnet-cvpr-2016} and DenseNet~\cite{densenet-cvpr-2107}. Note that the error rates are computed for the $top-1$ recognition prediction. In order to obtain a bounding box from a map, each approach uses a different threshold. For CAM~\cite{cam-cvpr-2016} and Grad-CAM~\cite{gradcam-iccv-2017} we used the threshold provided in the respective papers, for Backprop (for ResNet and DenseNet, other values from CAM) and cMWP~\cite{exbp-eccv-2016} we found the best performing thresholds on the same held out set. The values marked with $*$ for CAM are for a modified architecture where all $fc$ layers were replaced with a GAP layer and the model was retrained with the full ILSVRC training set ($1.2M$ images). Therefore, these numbers are not comparable. This is a limitation for CAM as it works only for networks with GAP layer and in modifying the architecture as explained above it loses recognition performance by $8.5\%$ and $2.2\%$ for AlexNet and VGG respectively.

\begin{figure*}[t]
\centering
   \includegraphics[width=\linewidth]{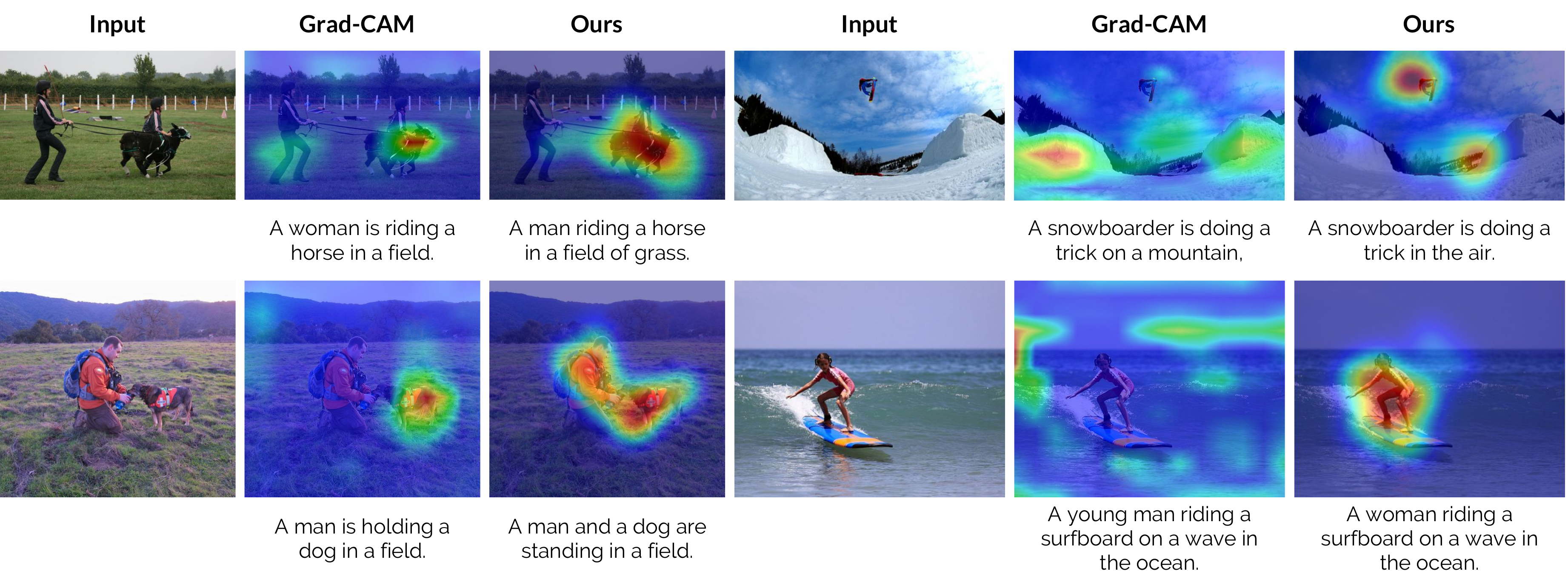}
\caption{Discriminative localization obtained by the proposed approach for captions predicted by the Show and Tell~\cite{SnT-pami-2016} model on sample images from MS COCO~\cite{mscoco-eccv-2014} dataset. Grad-CAM's illustrations are for Neuraltalk~\cite{neuraltalk-cvpr-2015} model. Note that the objects predicted in the captions are better highlighted for our method.}
\label{fig:captionMaps}
\end{figure*}

Figure \ref{fig:compareMaps} shows the comparison of maps between different approaches. The Table~\ref{tab:objLoc} shows that the proposed approach performs consistently well across a contrasting range of architectures, unlike other methods which perform well on selected architectures. Also, in Figure~\ref{fig:maps-at-intermediate} we present visualization at various layers in the architecture of VGG-16~\cite{vgg-iclr-2015}. Specifically, we show the evidence at all the five pool layers. Note that the fixations are computed on the feature maps which are of lower resolution compared to the input image. Therefore it is required to interpolate the location of fixations in order to show them on the input image. Observe that the localization improves as the resolution of the feature map increases, i.e., towards the input layer, fixations become more dense and accurate.

\subsection{Grounding Captions}
In this subsection, we show that our method can provide visual explanations for image captioning models. Caption generators predict a human readable sentence that describes contents of a given image. We present qualitative results for getting localization for the whole caption predicted by the Show and Tell~\cite{SnT-pami-2016} architecture.

The architecture has a CNN followed by an LSTM unit, which sequentially generates the caption word by word. The LSTM portion of the network is backtracked as discussed in section \ref{sec:lstm} following which we backtrack the CNN as discussed in sections {\ref{subsec:fc_text} and \ref{subsec:conv_text}}.

Figure~\ref{fig:captionMaps} shows the results where all the important objects that were predicted in the caption have been localized on the image. This shows that the proposed approach can effectively localize discriminative locations even for caption generators (i.e, grounding the caption). Our approach generalizes to deep neural networks trained for tasks other than object recognition. Note that some of the existing approaches discussed in the previous sections do not support localization for captions in their current version. For example, CAM~\cite{cam-cvpr-2016} requires a GAP layer in the model and Ex-BP~\cite{exbp-eccv-2016} grounds the tags predicted by a classification model instead of working with a caption generator.
\begin{figure*}[th]
\centering
\noindent\begin{minipage}{0.83\textwidth}
  \centering
  \begin{minipage}{.19\textwidth}
    \includegraphics[width=\linewidth]{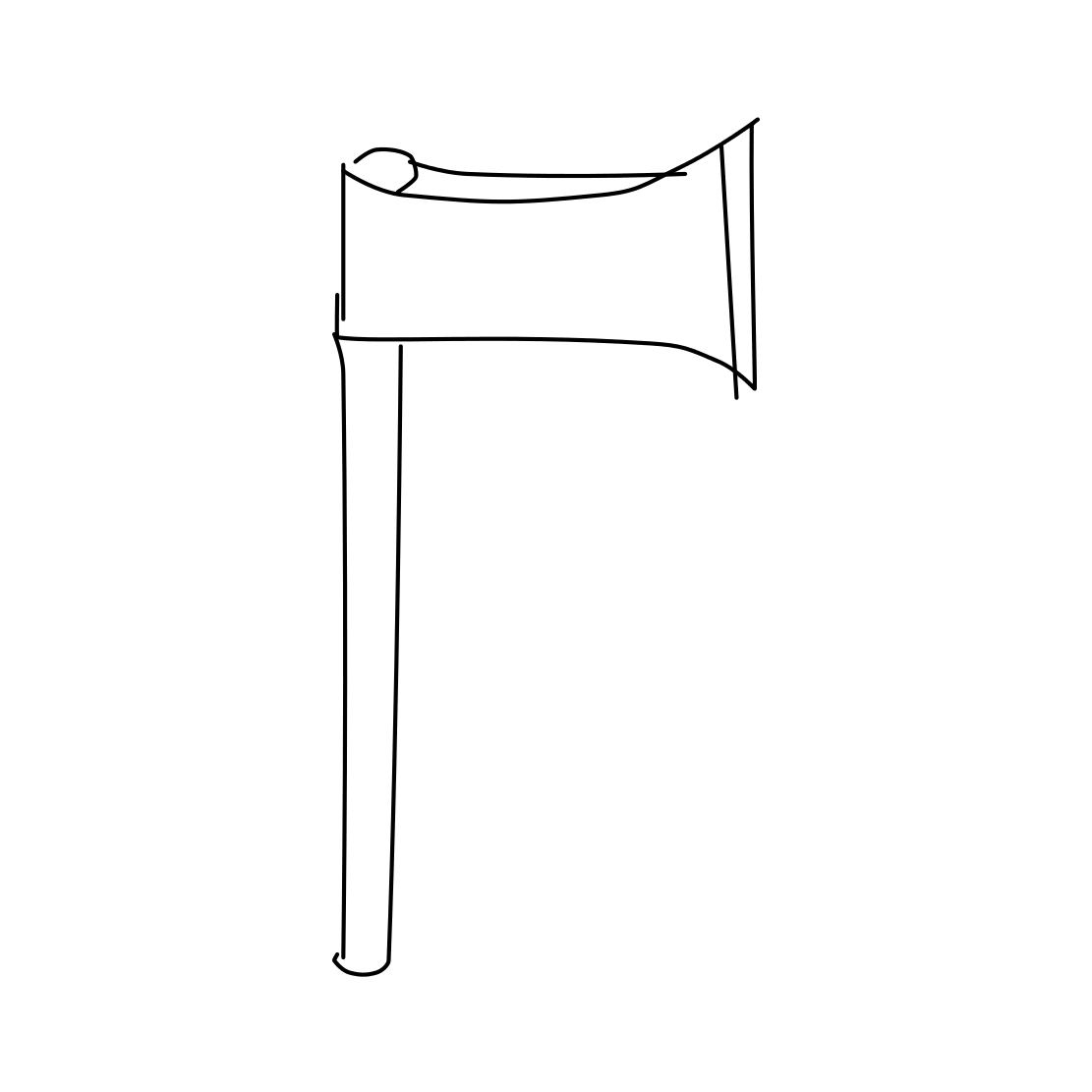}\caption*{Input Sketch}
  \end{minipage}
   \begin{minipage}{.19\textwidth}
    \includegraphics[width=\linewidth]{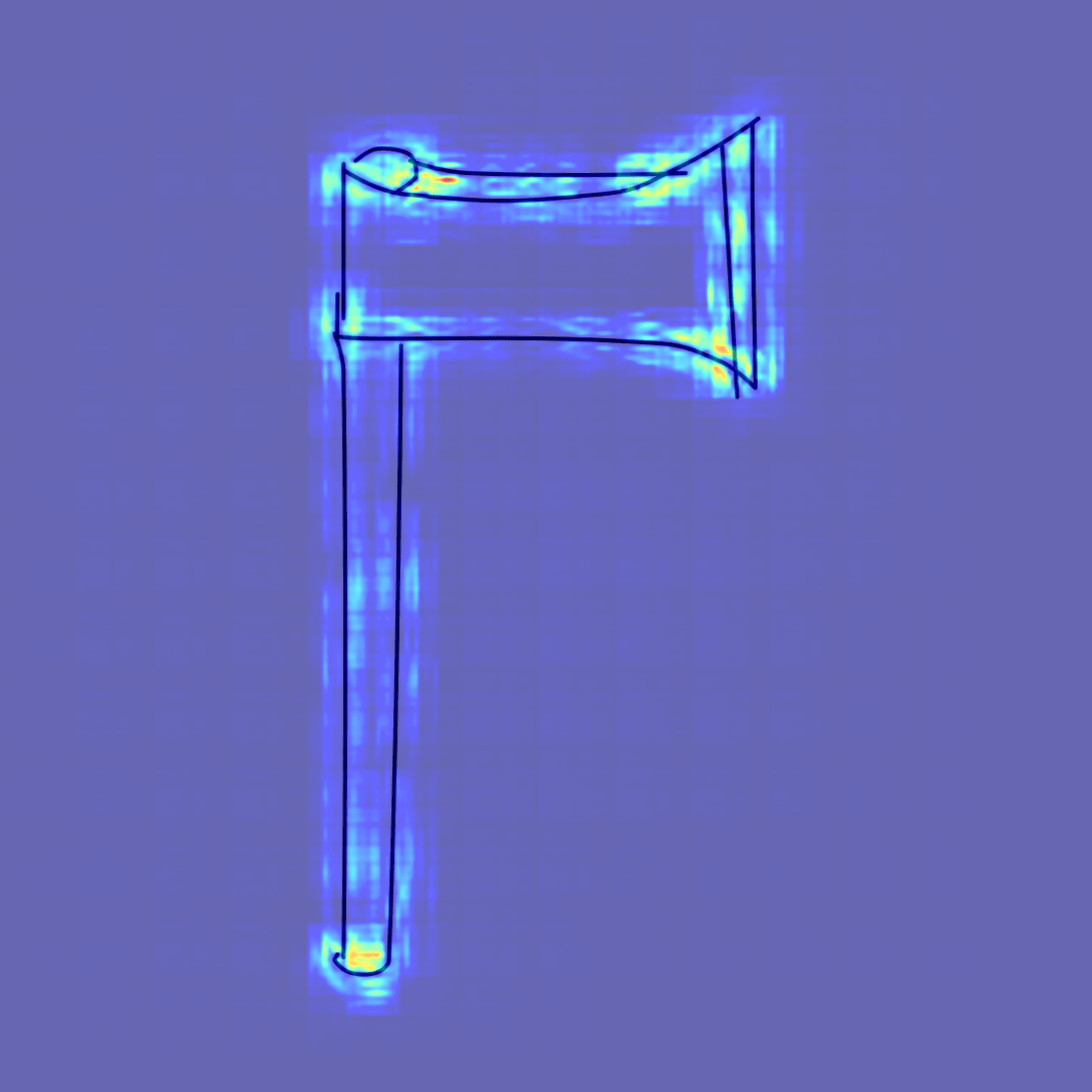}\caption*{BackProp~\cite{backprop-iclrw-2014}}
  \end{minipage}
  \begin{minipage}{.19\textwidth}
    \includegraphics[width=\linewidth]{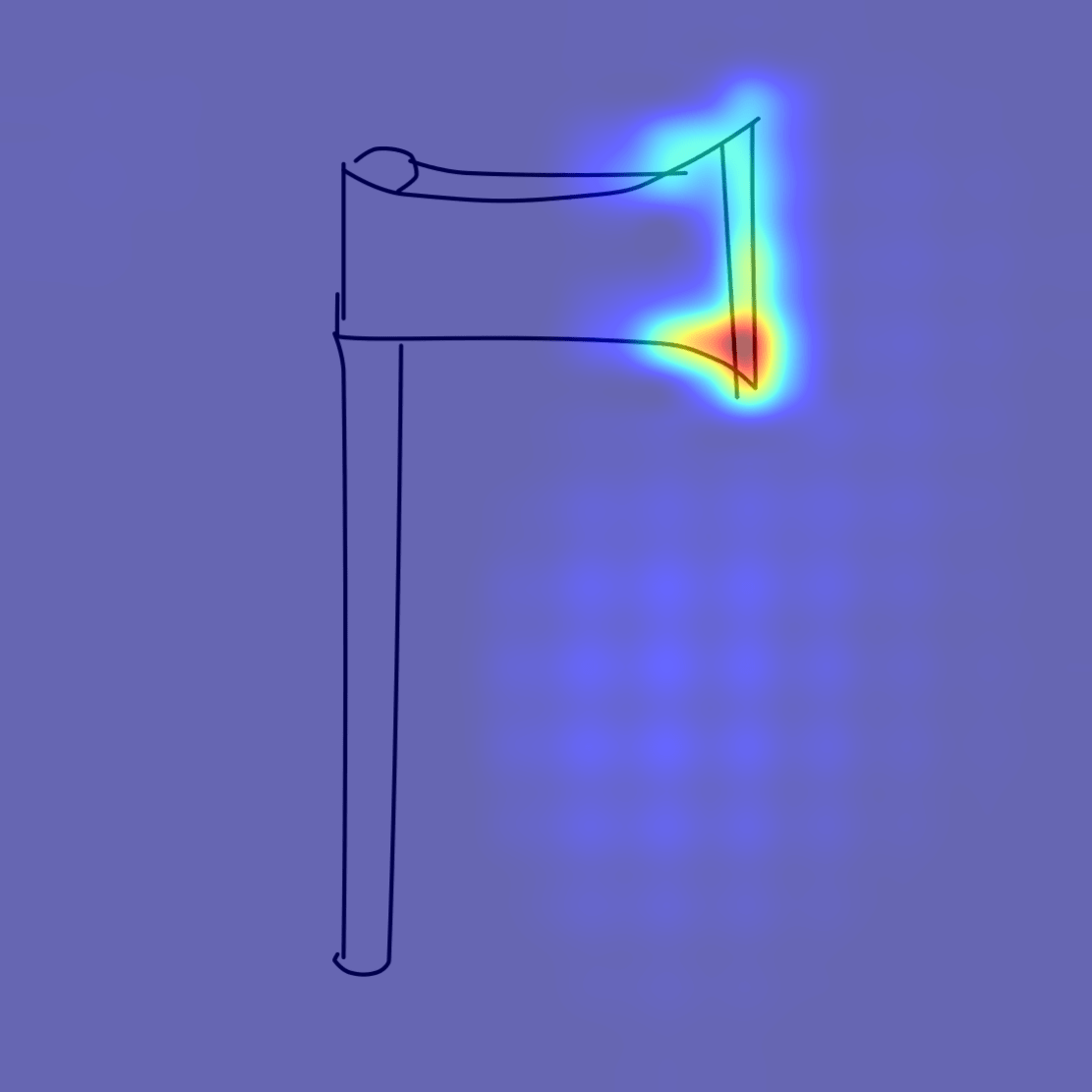}\caption*{cMWP~\cite{exbp-eccv-2016}}
  \end{minipage}
  \begin{minipage}{.19\textwidth}
    \includegraphics[width=\linewidth]{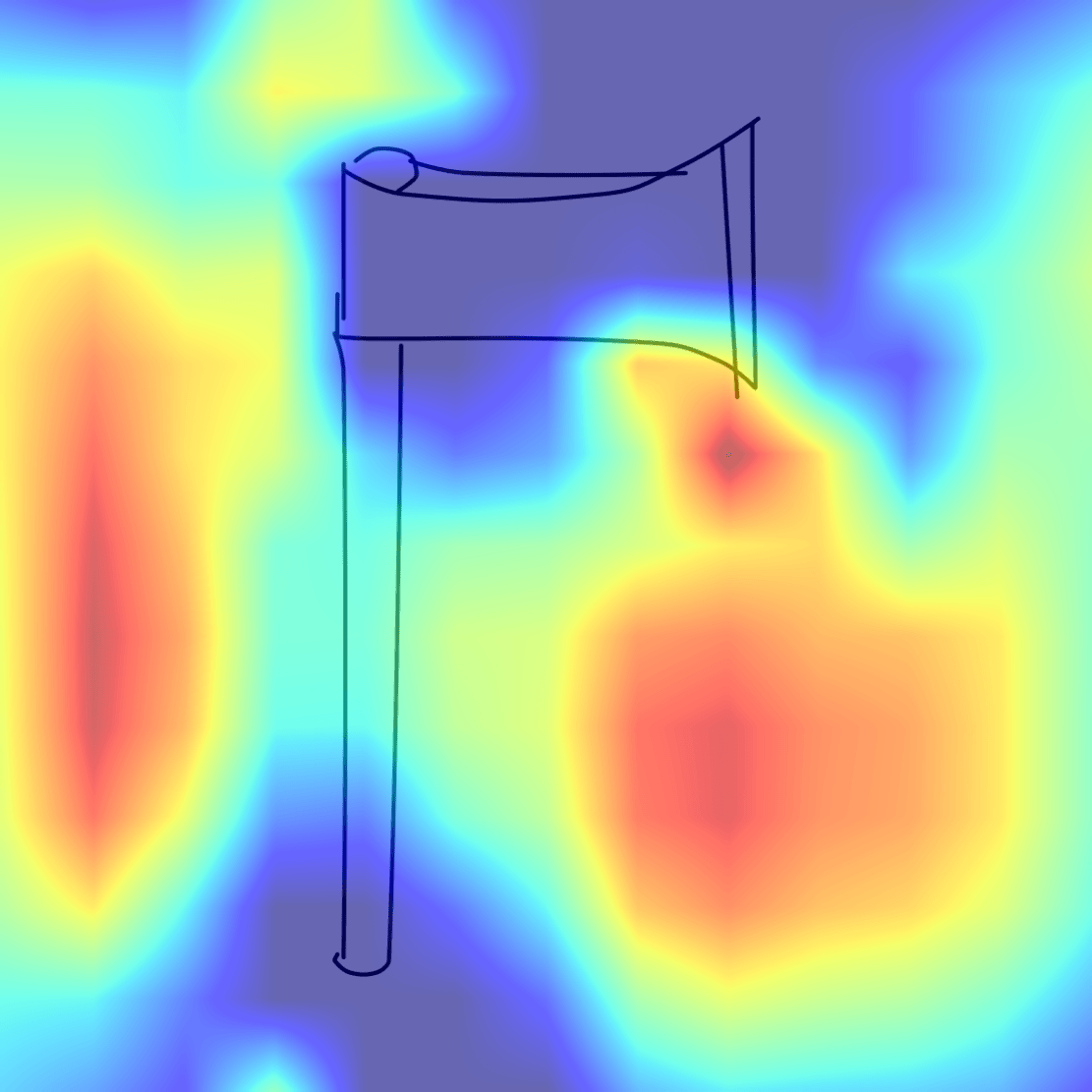}\caption*{Grad-CAM~\cite{gradcam-iccv-2017}}
  \end{minipage}
  \begin{minipage}{.19\textwidth}
    \includegraphics[width=\linewidth]{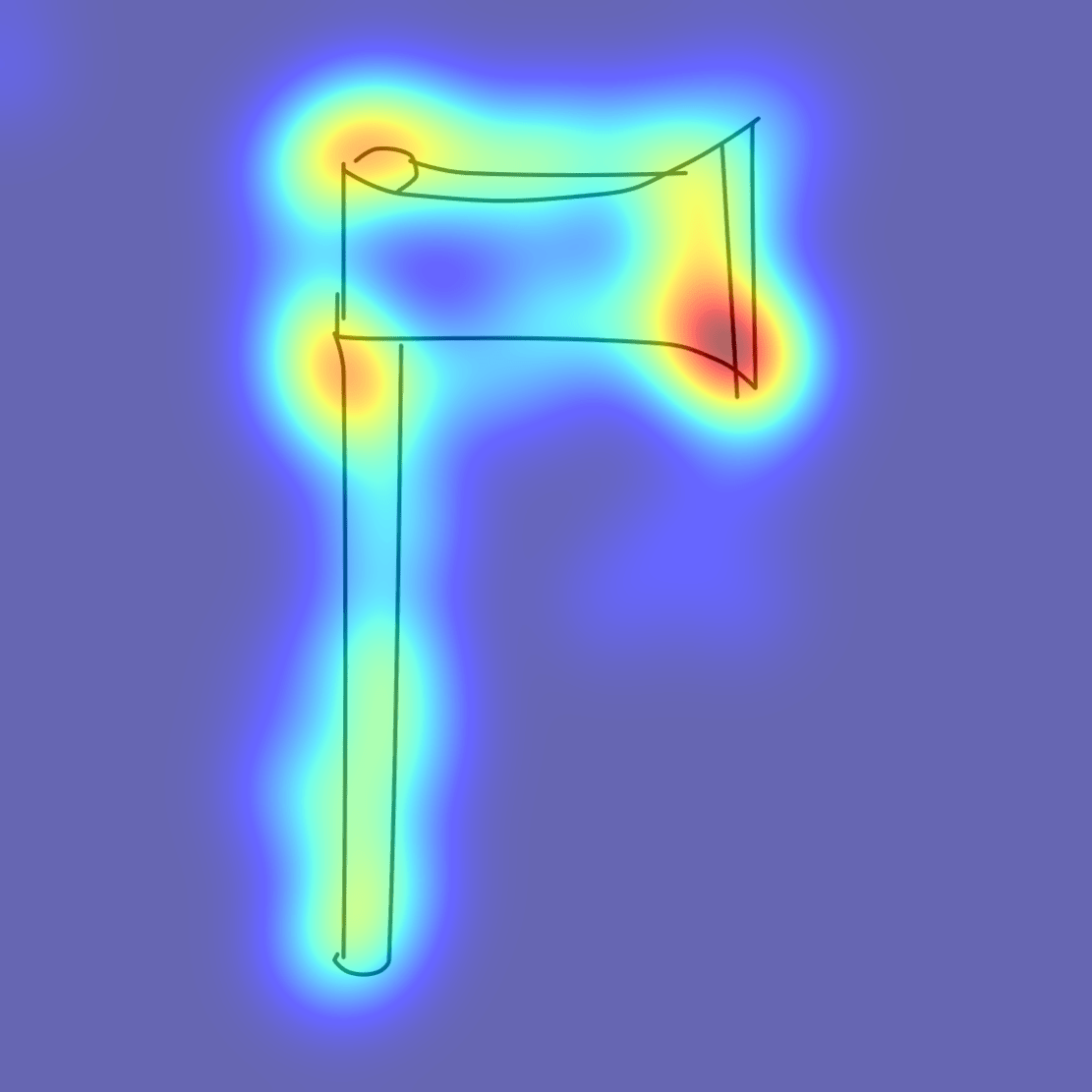}\caption*{Ours}
  \end{minipage}
  \vspace{0.005\textwidth}
\end{minipage}

\caption{Comparison of localization maps with different methods for a sketch classifier~\cite{onlinesketch-acmmm-2016}. }
\label{fig:sketchMaps}
\end{figure*}

\subsection{Saliency}
We now demonstrate the effectiveness of the proposed approach for predicting weakly-supervised saliency. The objective of this task is similar to that of Cholakkal \textit{et al.}~\cite{wssc-cvpr-2016}, where we perform weakly supervised saliency prediction using the models trained for object recognition. The ability of the proposed approach to provide visual explanations via back tracking the evidence onto the image is exploited for salient object detection.

Following \cite{wssc-cvpr-2016}, we perform the experiments on the Graz-2~\cite{graz2-cvpr-2007} dataset.
Graz-2 dataset consists of three classes namely bike, car and person. Each class has $150$ images for training and same number for testing. We fine-tuned VGG-16 architecture for recognizing these $3$ classes by replacing the final layer with $3$ units. We evaluated all approaches discussed in section \ref{subsec:objLoc} in addition to~\cite{wssc-cvpr-2016}. In order to obtain the saliency map from the fixations, we perform simple Gaussian blurring on the obtained CNN fixations. All the maps were thresholded based on the best thresholds we found on the train set for each approach. The evaluation is based on pixel-wise precision at equal error rate (EER) with the ground truth maps.

Table \ref{tab:sal} presents the precision rates per class for the Graz-2 dataset. Note that CAM~\cite{cam-cvpr-2016} was excluded as it does not work with the vanilla VGG~\cite{vgg-iclr-2015} network. This application highlights that the approaches which obtain maps at a low resolution and up-sample them to image size perform badly in this case due to the pixel level evaluation. However, our approach outperforms other methods to localize salient image regions by a huge margin.
\begin{table}[h]
\centering
\caption{Performance of different visualization methods for predicting saliency on Graz-2 dataset. Numbers denote the Pixel-wise precision at EER.}
\label{tab:sal}
\begin{tabular}{ccccc}

\hline
\textbf{Method} & \textbf{Bike}  & \textbf{Car}   & \textbf{Person} & \textbf{Mean}  \\ \hline
Backprop        & 39.51          & 28.50          & 42.64           & 36.88          \\
cMWP            & 61.84          & 46.82          & 44.02           & 50.89          \\
Grad-CAM        & 65.70          & {\ul 56.58}    & {\ul 57.98}     & 60.09          \\
WS-SC~\cite{wssc-cvpr-2016}           & {\ul 67.5}     & 56.48          & 57.56           & {\ul 60.52}    \\ \hline
Ours            & \textbf{71.21} & \textbf{62.15} & \textbf{61.27}  & \textbf{64.88} \\ \hline

\end{tabular}
\end{table}

\subsection{Localization across modalities}
We demonstrate that the proposed approach visualizes classifiers learned on other modalities as well. We perform the proposed CNN Fixations approach to show visualizations for a sketch classifier from~\cite{onlinesketch-acmmm-2016}. Sketches are a very different data modality compared to images. They are very sparse depictions of objects with only edges. CNNs trained to perform recognition on images are fine-tuned~\cite{onlinesketch-acmmm-2016,sketchanet-bmvc-2015} to perform recognition on sketches. We have considered AlexNet~\cite{deepcnn-nips-2012} fine-tuned over $160$  categories of sketches from Eitz dataset~\cite{eitz-atg-2012} to visualize the predictions.

Figure \ref{fig:sketchMaps} shows the localization maps for different approaches. We can clearly observe that the proposed approach highlights all the edges present in the sketches. This shows that our approach effectively localizes the sketches much better than the compared methods, showing it generalizes across different data modalities.

\begin{figure}[th]
\centering
\noindent\begin{minipage}{0.5\textwidth}
\captionsetup{justification=centering}
  \centering
  \begin{minipage}{.24\textwidth}
  	\centering
    \includegraphics[width=\linewidth]{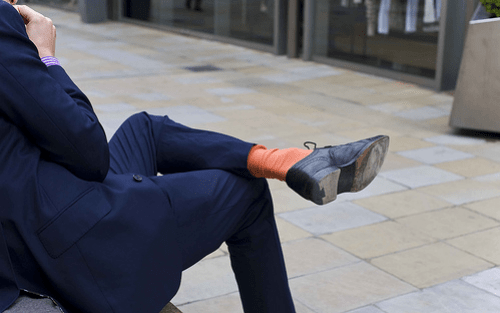}
    \caption*{{\color{green}Suit}}
  \end{minipage}
   \begin{minipage}{.24\textwidth}
   	\centering
    \includegraphics[width=\linewidth]{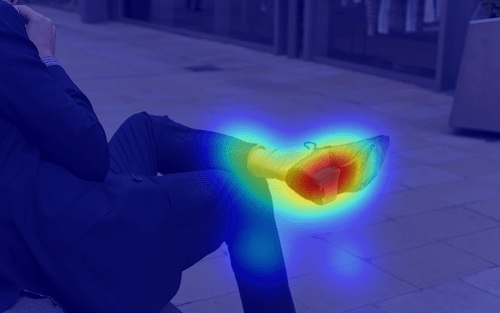}
    \caption*{{\color{red}Loafer}}
  \end{minipage}
  \begin{minipage}{.24\textwidth}
  	\centering
    \includegraphics[width=\linewidth]{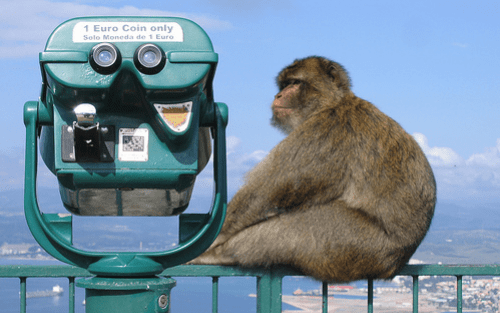}
    \caption*{{\color{green}Binoculars}}
  \end{minipage}
   \begin{minipage}{.24\textwidth}
   	\centering
    \includegraphics[width=\linewidth]{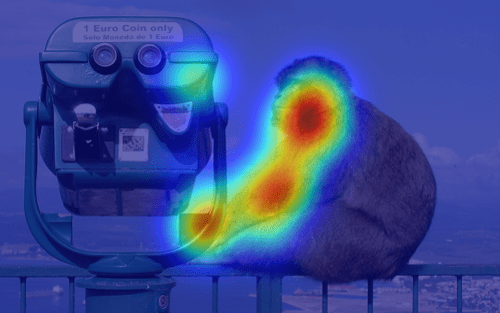}
     \caption*{{\color{red}Macaque}}
  \end{minipage}
  \vspace{0.002\textwidth}
\end{minipage}
\noindent\begin{minipage}{0.5\textwidth}
\captionsetup{justification=centering}
  \centering
  \begin{minipage}{.24\textwidth}
    \includegraphics[width=\linewidth]{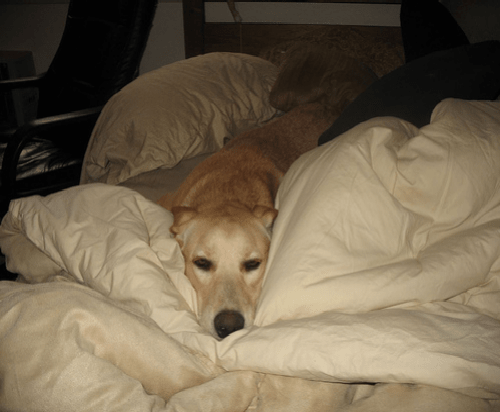}
    \caption*{{\color{green}Quilt / \\Comfortor}}
  \end{minipage}
   \begin{minipage}{.24\textwidth}
    \includegraphics[width=\linewidth]{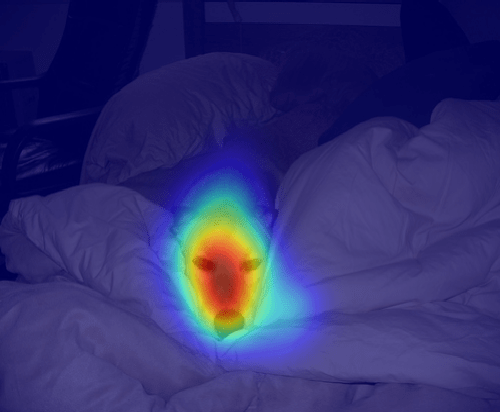}
    \caption*{{\color{red}Labrador\\Retriever}}
  \end{minipage}
  \begin{minipage}{.24\textwidth}
  	\centering
    \includegraphics[width=\linewidth]{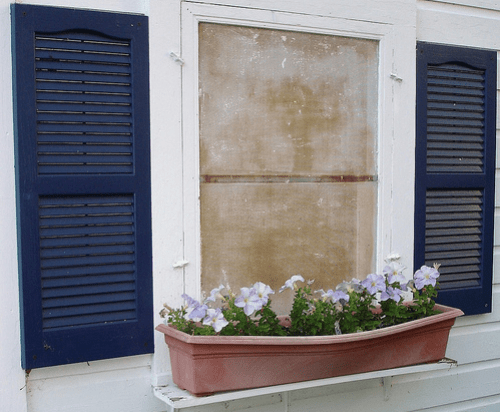}
    \caption*{{\color{green}Window\\Screen}}
  \end{minipage}
   \begin{minipage}{.24\textwidth}
   	\centering
    \includegraphics[width=\linewidth]{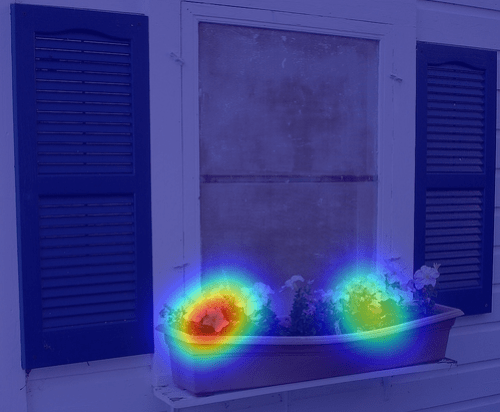}
    \caption*{{\color{red}Flower\\Pot}}
  \end{minipage}
    \vspace{0.002\textwidth}
\end{minipage}
\caption{Explaining the wrong recognition results obtained by VGG~\cite{vgg-iclr-2015}. Each pair of images along the rows show the image and its corresponding fixation map. Ground truth label is shown in green and the predicted is shown in red. Fixations clearly provide the explanations corresponding to the predicted labels.}
\label{fig:wrongPreds}
\end{figure}
\subsection{Explanations for erroneous predictions by CNNs}
CNNs are complex machine learning models offering very little transparency to analyse their inferences. For example, in cases where they wrongly predict the object category, it is required to diagnose them in order to understand what went wrong. If they can offer a proper explanation for their predictions, it is possible to improve various aspects of training and performance. The proposed CNN-fixations can act as a tool to help analyse the training process of CNN models.
\vspace{-0.1cm}
We demonstrate this by analysing the misclassified instances for object recognition. In Figure~\ref{fig:wrongPreds} we show sample images from ILSVRC validation images that are wrongly classified by VGG~\cite{vgg-iclr-2015}.
Each image is associated with the density map computed by our approach. Below each image-and-map pair, the ground truth and predicted labels are displayed in green and red respectively. Multiple objects are present in each of these images. The CNN recognizes the objects that are not labeled but are present in the images. The computed maps for the predicted labels accurately locate those objects such as loafer, macaque, etc. and offer visual explanations for the CNN’s behaviour. It is evident that these images are labeled ambiguously and the proposed method can help improve the annotation quality of the data.
\subsection{Presence of Adversarial noise}
\begin{figure}[t]
\centering
\noindent\begin{minipage}{0.5\textwidth}
  \centering
  \begin{minipage}{.19\textwidth}
    \includegraphics[width=\linewidth]{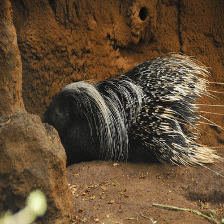}
    \caption*{{\color{green}Porcupine}}
  \end{minipage}
  \begin{minipage}{.19\textwidth}
    \includegraphics[width=\linewidth]{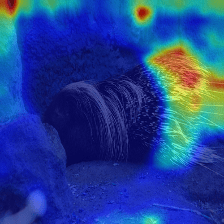}
    \caption*{}
  \end{minipage}
  \begin{minipage}{.19\textwidth}
    \includegraphics[width=\linewidth]{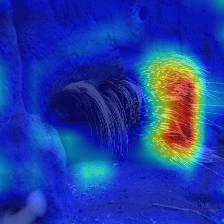}
    \caption*{}
  \end{minipage}
  \begin{minipage}{.19\textwidth}
    \includegraphics[width=\linewidth]{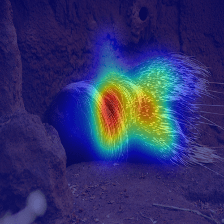}
    \caption*{}
  \end{minipage}
  \vspace{0.005\textwidth}
\end{minipage}
\noindent\begin{minipage}{0.5\textwidth}
  \centering
  \begin{minipage}{.19\textwidth}
    \includegraphics[width=\linewidth]{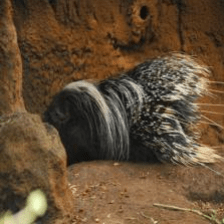}
    \caption*{{\color{red}Marmoset}}
  \end{minipage}
  \begin{minipage}{.19\textwidth}
    \includegraphics[width=\linewidth]{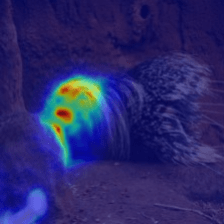}
    \caption*{}
  \end{minipage}
  \begin{minipage}{.19\textwidth}
    \includegraphics[width=\linewidth]{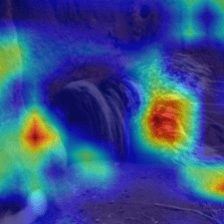}
    \caption*{}
  \end{minipage}
  \begin{minipage}{.19\textwidth}
    \includegraphics[width=\linewidth]{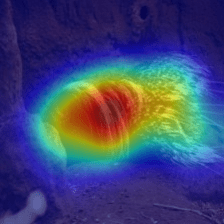}
    \caption*{}
  \end{minipage}
  \vspace{0.005\textwidth}
\end{minipage}
\noindent\begin{minipage}{0.5\textwidth}
  \centering
  \begin{minipage}{.19\textwidth}
    \includegraphics[width=\linewidth]{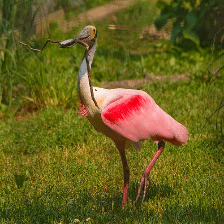}
    \caption*{{\color{green}Spoonbill}}
  \end{minipage}
  \begin{minipage}{.19\textwidth}
    \includegraphics[width=\linewidth]{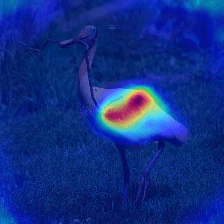}
    \caption*{}
  \end{minipage}
  \begin{minipage}{.19\textwidth}
    \includegraphics[width=\linewidth]{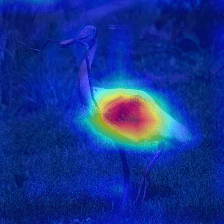}
    \caption*{}
  \end{minipage}
  \begin{minipage}{.19\textwidth}
    \includegraphics[width=\linewidth]{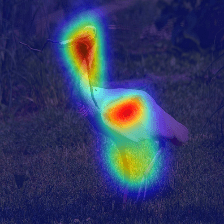}
    \caption*{}
  \end{minipage}
  \vspace{0.005\textwidth}
\end{minipage}
\noindent\begin{minipage}{0.5\textwidth}
  \centering
  \begin{minipage}{.19\textwidth}
  	\centering
    \includegraphics[width=\linewidth]{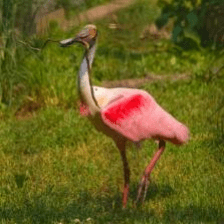}
    \caption*{{\color{red}Crane}}
  \end{minipage}
  \begin{minipage}{.19\textwidth}
  	\centering
    \includegraphics[width=\linewidth]{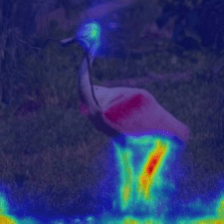}
    \caption*{cMWP~\cite{exbp-eccv-2016}}
  \end{minipage}
  \begin{minipage}{.19\textwidth}
  	\centering
    \includegraphics[width=\linewidth]{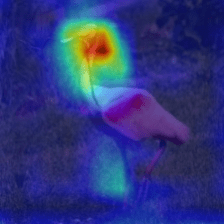}
    \caption*{GCAM~\cite{gradcam-iccv-2017}}
  \end{minipage}
  \begin{minipage}{.19\textwidth}
  	\centering
    \includegraphics[width=\linewidth]{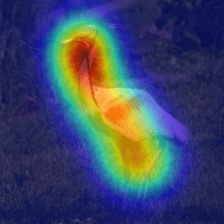}
    \caption*{Ours}
  \end{minipage}
  \vspace{0.002\textwidth}
\end{minipage}
\caption{Visual explanations for sample adversarial images provided by multiple methods. First and third rows show the evidence for the clean samples for which the predicted label is shown in green. Second and fourth rows show the same for the corresponding DeepFool~\cite{deepfool-cvpr-2016} adversaries for which the label is shown in red.}
\label{fig:advMaps}
\end{figure}
Many recent works~(e.g. \cite{deepfool-cvpr-2016, explainingharnessing-iclr-2015,mopuri-bmvc-2017}) have demonstrated the susceptibility of convolutional neural networks to \emph{Adversarial} samples. These are images that have been perturbed with structured quasi-imperceptible noise towards the objective of fooling the classifier. Figure~\ref{fig:advMaps} shows two samples of such images that have been perturbed using the DeepFool method~\cite{deepfool-cvpr-2016} for the VGG-16 network. 
The figure clearly shows that even though the label is changed by the added perturbation, the proposed approach is still able to correctly localize the object regions in both cases. Note that the explanations provided by the gradient based methods~(e.g. \cite{gradcam-iccv-2017}) get affected by the adversarial perturbation. This shows that our approach is robust to images perturbed with adversarial noise to locate the object present in the image.
\subsection{Generic Object Proposal}
In this subsection we demonstrate that CNNs trained for object recognition can also act as generic object detectors. Existing algorithms for this task (e.g.~\cite{selsearch-ijcv-2013,bing-cvpr-2014,eboxes-eccv-2014,mcg-cvpr-2014}) typically provide hundreds of class agnostic proposals, and their performance is evaluated by the average precision and recall measures. While most of them perform very well for large number of proposals, it is more useful to get better metrics at lower number of proposals. Investigating the performances for thousands of proposals is not appropriate since a typical image rarely contains more than a handful of objects. Recent approaches (e.g. \cite{craft-cvpr-2016}) attempt for achieving better performances at smaller number of proposals. In this section, we take this notion to extreme and investigate the performance of these approaches for \emph{\textbf{single}} best proposal. This is because, the proposed method can provide visual explanation for the predicted label and while doing so it can locate the object region using a single proposal. Therefore it is fair to compare our proposal with the best proposal of multiple region proposal algorithms.

Using the proposed approach, we generate object proposals for unseen object categories. We evaluated the models trained over ILSVRC dataset on the PASCAL VOC-$2007$~\cite{pascal-voc-2007} test images. Note that the target categories are different from that of the training dataset and the models are trained for object recognition. We pass each image in the test set through the CNN and obtain a bounding box (for the predicted label) as explained in \ref{subsec:objLoc}. This proposal is compared with the ground truth bounding box of the image and if the IoU is more than $0.5$, it is considered a true positive. We then measure the performance in terms of the mean average recall and precision per class as done in the PASCAL benchmark \cite{pascal-voc-2007} and \cite{stl-wacv-2016}.

Table~\ref{tab:proposals-voc} shows the performance of the proposed approach for single proposal and compares it against well known object proposal approaches and other CNN based visualization methods discussed above. For STL~\cite{stl-wacv-2016} the numbers were obtained from their paper and for other CNN based approaches we used GoogLeNet~\cite{googlenet-cvpr-21014} as the underlying CNN. The objective of this experiment is to demonstrate the ability of CNNs as generic object detectors via localizing evidence for the prediction. The proposed approach outperforms all the non-CNN based methods by large margin and performs better than all the CNN based methods except the Backprop~\cite{backprop-iclrw-2014} and DeepMask~\cite{deepmask-nips-2015} methods, which perform equally. Note that~\cite{deepmask-nips-2015}, in spite of using a strong net (ResNet) and training procedure to predict a class agnostic segmentation, performs comparable to our method.

\begin{table}[t]
\centering
\caption{The performance of different methods for Generic Object Proposal generation on the PASCAL VOC-$2007$ test set. Note that the methods are divided into CNN based and non-CNN based also the proposed method outperforms all the methods along with backprop~\cite{backprop-iclrw-2014} method. All the CNN based works except \cite{deepmask-nips-2015} use the GoogLeNet~\cite{googlenet-cvpr-21014} and \cite{deepcnn-nips-2012} uses a ResNet~\cite{resnet-cvpr-2016} architecture to compute the metrics. In spite of working with the best CNN, \cite{deepmask-nips-2015} performs on par with our approach (denoted with $*$).}
\label{tab:proposals-voc}
\begin{tabular}{c|lcc}
\hline
\multicolumn{1}{l|}{\textbf{Type}} & \textbf{Method}                               & \multicolumn{1}{l}{\textbf{mRecall}} & \multicolumn{1}{l}{\textbf{mPrecision}} \\ \hline
\multirow{4}{*}{Non-CNN}  & \multicolumn{1}{c}{Selective Search} & 0.10                        & 0.14                           \\
                          & EdgeBoxes                            & 0.18                        & 0.26                           \\
                          & MCG                                  & 0.17                        & 0.25                           \\
                          & BING                                 & 0.18                        & 0.25                           \\ \hline
\multirow{7}{*}{CNN}      & Backprop                             & \textbf{0.32}                        & \uline{0.36}                           \\
                          & CAM                                  & \uline{0.30}                        & 0.33                           \\
                          & cMWP                                 & 0.23                        & 0.26                           \\
                          & Grad-CAM                             & 0.18                        & 0.21                           \\
                          & STL-WL                               & 0.23                        & 0.31                           \\ 
                          & Deep Mask~\cite{deepmask-nips-2015}                    & 0.29*                        & \textbf{0.38*}                           \\ \cline{2-4} 
                          & Ours                                 & \textbf{0.32}                        & \uline{0.36}                           \\ \hline 
\end{tabular}
\end{table}

\section{Conclusion}
\label{sec:conclu}
We propose an unfolding approach to trace the evidence
for a given neuron activation, in the preceding layers. Based
on this, a novel visualization technique, CNN-fixations is
presented to highlight the image locations that are responsible for the predicted label. High resolution and discriminative localization maps are computed from these locations.
The proposed approach is computationally very efficient which unlike other existing approaches doesn't require to compute either the gradients or the prediction differences. Our method effectively exploits the feature dependencies that evolve out of the end-to-end training process. As a result only a single forward pass is sufficient to provide a faithful visual explanation for the predicted label. 

We also demonstrate that our approach enables interesting set of applications. Furthermore, in cases of erroneous predictions, the proposed approach offers visual explanations to make the CNN models more transparent and help improve the training process and annotation procedure.
\section{Acknowledgements}
The authors thank Suraj Srinivas for having helpful discussions while conducting this research.
\bibliographystyle{ieee}
\bibliography{mybibliography}

\begin{thebibliography}{10}\itemsep=-1pt

\bibitem{tensorflow2015-whitepaper}
M.~Abadi~{et~al.}
\newblock {TensorFlow}: Large-scale machine learning on heterogeneous systems,
  2015.

\bibitem{mcg-cvpr-2014}
P.~Arbel\'{a}ez, J.~Pont-Tuset, J.~Barron, F.~Marques, and J.~Malik.
\newblock Multiscale combinatorial grouping.
\newblock In {\em IEEE Computer Vision and Pattern Recognition, {(CVPR)}},
  2014.

\bibitem{layerwiserelevance-plos-2015}
S.~Bach, A.~Binder, G.~Montavon, F.~Klauschen, K.-R. M{\"u}ller, and W.~Samek.
\newblock On pixel-wise explanations for non-linear classifier decisions by
  layer-wise relevance propagation.
\newblock {\em PloS one}, 10(7), 2015.

\bibitem{stl-wacv-2016}
L.~Bazzani, A.~Bergamo, D.~Anguelov, and L.~Torresani.
\newblock Self-taught object localization with deep networks.
\newblock In {\em IEEE Winter Conference on Applications of Computer Vision
  (WACV)}, 2016.

\bibitem{bing-cvpr-2014}
M.~M. Cheng, Z.~Zhang, W.~Y. Lin, and P.~H.~S. Torr.
\newblock {BING}: Binarized normed gradients for objectness estimation at
  300fps.
\newblock In {\em IEEE Computer Vision and Pattern Recognition {(CVPR)}}, 2014.

\bibitem{wssc-cvpr-2016}
H.~Cholakkal, J.~Johnson, and D.~Rajan.
\newblock Backtracking {S}c{SPM} image classifier for weakly supervised
  top-down saliency.
\newblock In {\em IEEE Conference on Computer Vision and Pattern Recognition
  (CVPR)}, 2016.

\bibitem{eitz-atg-2012}
M.~Eitz, J.~Hays, and M.~Alexa.
\newblock How do humans sketch objects?
\newblock {\em ACM Transactions on Graphics (TOG)}, 31(4), 2012.

\bibitem{pascal-voc-2007}
M.~Everingham, L.~Van~Gool, C.~K.~I. Williams, J.~Winn, and A.~Zisserman.
\newblock The {PASCAL} {V}isual {O}bject {C}lasses {C}hallenge 2007 {(VOC2007)}
  {R}esults.

\bibitem{explainingharnessing-iclr-2015}
I.~J. Goodfellow, J.~Shlens, and C.~Szegedy.
\newblock Explaining and harnessing adversarial examples.
\newblock In {\em International Conference on Learning Representations (ICLR)},
  2015.

\bibitem{resnet-cvpr-2016}
K.~He, X.~Zhang, S.~Ren, and J.~Sun.
\newblock Deep residual learning for image recognition.
\newblock In {\em IEEE conference on computer vision and pattern recognition
  (CVPR)}, 2016.

\bibitem{lstm-nc-1997}
S.~Hochreiter and J.~Schmidhuber.
\newblock Long short-term memory.
\newblock {\em Neural computation}, 9(8), 1997.

\bibitem{densenet-cvpr-2107}
G.~Huang and Z.~Liu.
\newblock Densely connected convolutional networks.
\newblock In {\em IEEE Conference on Computer Vision and Pattern Recognition
  (CVPR)}, 2017.

\bibitem{caffe-acmmm-2014}
Y.~Jia, E.~Shelhamer, J.~Donahue, S.~Karayev, J.~Long, R.~Girshick,
  S.~Guadarrama, and T.~Darrell.
\newblock Caffe: Convolutional architecture for fast feature embedding.
\newblock In {\em Proceedings of the ACM International Conference on
  Multimedia}, 2014.

\bibitem{neuraltalk-cvpr-2015}
A.~Karpathy and F.~Li.
\newblock Deep visual-semantic alignments for generating image descriptions.
\newblock In {\em IEEE Conference on Computer Vision and Pattern Recognition
  {(CVPR)}}, 2015.

\bibitem{deepcnn-nips-2012}
A.~Krizhevsky, I.~Sutskever, and G.~E. Hinton.
\newblock Imagenet classification with deep convolutional neural networks.
\newblock In {\em Advances inNeural Information Processing Systems (NIPS)}.
  2012.

\bibitem{mscoco-eccv-2014}
T.-Y. Lin, M.~Maire, S.~Belongie, J.~Hays, P.~Perona, D.~Ramanan,
  P.~Doll{\'a}r, and C.~L. Zitnick.
\newblock Microsoft coco: Common objects in context.
\newblock In {\em European conference on computer vision (ECCV)}, 2014.

\bibitem{graz2-cvpr-2007}
M.~Marszalek and C.~Schmid.
\newblock Accurate object localization with shape masks.
\newblock In {\em IEEE Conference on Computer Vision and Pattern Recognition
  {(CVPR)}}, 2007.

\bibitem{deepfool-cvpr-2016}
S.~Moosavi{-}Dezfooli, A.~Fawzi, and P.~Frossard.
\newblock Deepfool: {A} simple and accurate method to fool deep neural
  networks.
\newblock In {\em IEEE Conference on Computer Vision and Pattern Recognition
  {(CVPR)}}, 2016.

\bibitem{mopuri-bmvc-2017}
K.~R. Mopuri, U.~Garg, and R.~V. Babu.
\newblock Fast feature fool: A data independent approach to universal
  adversarial perturbations.
\newblock In {\em Proceedings of the British Machine Vision Conference
  ({BMVC})}, 2017.

\bibitem{deepmask-nips-2015}
P.~O. Pinheiro, R.~Collobert, and P.~Doll\'ar.
\newblock Learning to segment object candidates.
\newblock In {\em Advances in Neural Information Processing Systems (NIPS)},
  2015.

\bibitem{fasterrcnn-nips-2015}
S.~Ren, K.~He, R.~Girshick, and J.~Sun.
\newblock Faster r-cnn: Towards real-time object detection with region proposal
  networks.
\newblock In {\em Advances in Neural Information Processing Systems (NIPS)}.
  2015.

\bibitem{relevance-kde-2008}
M.~Robnik-{\v{S}}ikonja and I.~Kononenko.
\newblock Explaining classifications for individual instances.
\newblock {\em IEEE Transactions on Knowledge and Data Engineering}, 20(5),
  2008.

\bibitem{imagenet-ijcv-2015}
O.~Russakovsky, J.~Deng, H.~Su, J.~Krause, S.~Satheesh, S.~Ma, Z.~Huang,
  A.~Karpathy, A.~Khosla, M.~Bernstein, A.~C. Berg, and L.~Fei-Fei.
\newblock {ImageNet Large Scale Visual Recognition Challenge}.
\newblock {\em International Journal of Computer Vision (IJCV)}, 115(3), 2015.

\bibitem{onlinesketch-acmmm-2016}
R.~K. Sarvadevabhatla, J.~Kundu, and R.~V. Babu.
\newblock Enabling my robot to play pictionary: Recurrent neural networks for
  sketch recognition.
\newblock In {\em ACM Conference on Multimedia}, 2016.

\bibitem{gradcam-iccv-2017}
R.~R. Selvaraju, M.~Cogswell, A.~Das, R.~Vedantam, D.~Parikh, and D.~Batra.
\newblock Grad-cam: Visual explanations from deep networks via gradient-based
  localization.
\newblock In {\em The IEEE International Conference on Computer Vision (ICCV)},
  2017.

\bibitem{ohem-cvpr-2016}
A.~Shrivastava, A.~Gupta, and R.~Girshick.
\newblock Training region-based object detectors with online hard example
  mining.
\newblock In {\em IEEE Conference on Computer Vision and Pattern Recognition
  (CVPR)}, 2016.

\bibitem{backprop-iclrw-2014}
K.~Simonyan, A.~Vedaldi, and A.~Zisserman.
\newblock Deep inside convolutional networks: Visualising image classification
  models and saliency maps.
\newblock In {\em International Conference on Learning Representations {(ICLR)}
  Workshops}, 2014.

\bibitem{vgg-iclr-2015}
K.~Simonyan and A.~Zisserman.
\newblock Very deep convolutional networks for large-scale image recognition.
\newblock In {\em International Conference on Learning Representations
  ({ICLR})}, 2015.

\bibitem{guidedbackprop-iclrw-2015}
J.~Springenberg, A.~Dosovitskiy, T.~Brox, and M.~Riedmiller.
\newblock Striving for simplicity: The all convolutional net.
\newblock In {\em International Conference on Learning Representations (ICLR)
  (workshop track)}, 2015.

\bibitem{googlenet-cvpr-21014}
C.~Szegedy, W.~Liu, Y.~Jia, P.~Sermanet, S.~Reed, D.~Anguelov, D.~Erhan,
  V.~Vanhoucke, and A.~Rabinovich.
\newblock Going deeper with convolutions.
\newblock In {\em IEEE conference on computer vision and pattern recognition
  (CVPR)}, 2015.

\bibitem{selsearch-ijcv-2013}
J.~R. Uijlings, K.~E. Van De~Sande, T.~Gevers, and A.~W. Smeulders.
\newblock Selective search for object recognition.
\newblock {\em International journal of computer vision (IJCV)}, 104(2), 2013.

\bibitem{SnT-pami-2016}
O.~Vinyals, A.~Toshev, S.~Bengio, and D.~Erhan.
\newblock Show and tell: Lessons learned from the 2015 mscoco image captioning
  challenge.
\newblock {\em IEEE transactions on pattern analysis and machine intelligence
  (TPAMI)}, 39(4), 2017.

\bibitem{craft-cvpr-2016}
B.~Yang, J.~Yan, Z.~Lei, and S.~Li.
\newblock Craft objects from images.
\newblock In {\em IEEE Conference on Computer Vision and Pattern Recognition
  (CVPR)}, 2016.

\bibitem{sketchanet-bmvc-2015}
Q.~Yu, Y.~Yang, Y.~Song, T.~Xiang, and T.~M. Hospedales.
\newblock Sketch-a-net that beats humans.
\newblock In {\em British Machine Vision Conference {(BMVC)}}, 2015.

\bibitem{deconv-eccv-2014}
M.~D. Zeiler and R.~Fergus.
\newblock Visualizing and understanding convolutional networks.
\newblock In {\em European Conference on Computer Vision {(ECCV)}}, 2014.

\bibitem{exbp-eccv-2016}
J.~Zhang, Z.~Lin, S.~X. Brandt, Jonathan, and S.~Sclaroff.
\newblock Top-down neural attention by excitation backprop.
\newblock In {\em European Conference on Computer Vision (ECCV)}, 2016.

\bibitem{cam-cvpr-2016}
B.~Zhou, A.~Khosla, A.~Lapedriza, A.~Oliva, and A.~Torralba.
\newblock Learning deep features for discriminative localization.
\newblock In {\em IEEE Computer Vision and Pattern Recognition (CVPR)}, 2016.

\bibitem{preddiff-iclr-2017}
L.~M. Zintgraf, T.~S. Cohen, T.~Adel, and M.~Welling.
\newblock Visualizing deep neural network decisions: Prediction difference
  analysis.
\newblock In {\em International Conference on Learning Representations
  {(ICLR)}}, 2017.

\bibitem{eboxes-eccv-2014}
C.~L. Zitnick and P.~Doll\'ar.
\newblock Edge boxes: Locating object proposals from edges.
\newblock In {\em European Conference on Computer Vision (ECCV)}, 2014.

\end{thebibliography}

\end{document}